\documentclass[preprint,12pt]{elsarticle}

\usepackage{amssymb}
\usepackage{multirow}
\usepackage{booktabs}
\usepackage{amsmath}
\usepackage{graphicx}
\usepackage{subfig}
\usepackage{autobreak}
\usepackage{url}
\usepackage{microtype}
\usepackage{longtable}

\journal{Pattern Recognition}

\begin{document}

\begin{frontmatter}



\title{AdvCloak: Customized Adversarial Cloak for Privacy Protection}


\author{Xuannan~Liu}
\author{Yaoyao~Zhong}
\author{Xing~Cui}
\author{Yuhang~Zhang}
% \author{Peipei~Li\corref{correspondingauthor}}
\author{\\Peipei~Li}
\author{Weihong~Deng}
% \author{Zhaofeng~He}
% \ead{zhaofenghe@bupt.edu.cn}

%\fntext[co-firstauthor]{These authors contributed equally to this work and should be considered co-first authors.}
% \cortext[correspondingauthor]{Corresponding author.}

%\affiliation{organization={School of Artificial Intelligence, Beijing University of Posts and Telecommunications},
%            city={Beijing},
%            postcode={100876}, 
%            country={China}}

\address{School of Artificial Intelligence, Beijing University of Posts and \\
	Telecommunications, Beijing, 100876, China}

\begin{abstract}
With extensive face images being shared on social media, there has been a notable escalation in privacy concerns. In this paper, we propose AdvCloak, an innovative framework for privacy protection using generative models. AdvCloak is designed to automatically customize class-wise adversarial masks that can maintain superior image-level naturalness while providing enhanced feature-level generalization ability. Specifically, AdvCloak sequentially optimizes the generative adversarial networks by employing a two-stage training strategy. This strategy initially focuses on adapting the masks to the unique individual faces via image-specific training and then enhances their feature-level generalization ability to diverse facial variations of individuals via person-specific training. To fully utilize the limited training data, we combine AdvCloak with several general geometric modeling methods, to better describe the feature subspace of source identities. Extensive quantitative and qualitative evaluations on both common and celebrity datasets demonstrate that AdvCloak outperforms existing state-of-the-art methods in terms of efficiency and effectiveness.
\end{abstract}

% \centering
% \includegraphics[width=1.0\linewidth]{Fig/pipline.jpg}

% \begin{highlights}
% \item We propose a novel privacy protection framework based on the generative models named AdvCloak, which can automatically customize imperceptible class-wise privacy masks for each individual.
% \item Two-stage training strategy is presented to crisply address the challenges of both the image-level visual naturalness and feature-level generalization ability.
% \item Extensive qualitative and quantitative experiments demonstrate that AdvCloak achieves state-of-the-art performance in the face privacy protection community with pleasing visual naturalness.
% \end{highlights}


\end{frontmatter}



\clearpage
\section{Introduction}
With the rapid development of deep learning, face recognition (FR)~\cite{taigman2014deepface,schroff2015facenet,liu2017sphereface,wang2018cosface,deng2019arcface,zhong2021sface} has achieved significant breakthroughs, leading to the widespread application in numerous fields~\cite{li2020deep, fang2020identity, lu2022transferring, xu2024depth}. However, the inappropriate use of this technology also raises serious concerns regarding personal privacy and security. For instance, unauthorized entities can maliciously scrape face images from social media (such as Twitter, Facebook, WeChat, etc.) and utilize FR systems to extract personal identity information without consent~\cite{shan2020fawkes}. Despite the stringent privacy protection laws (\textit{e.g.}, GDPR, CCPA, \textit{etc.})~\cite{privacylaw}, there persists the invasion of privacy information by malicious individuals. Consequently, there is an urgent need to explore effective strategies to help individuals protect their face images against unauthorized FR systems.

\begin{figure}[!ht]
	\centering
	\includegraphics[width=0.90\linewidth]{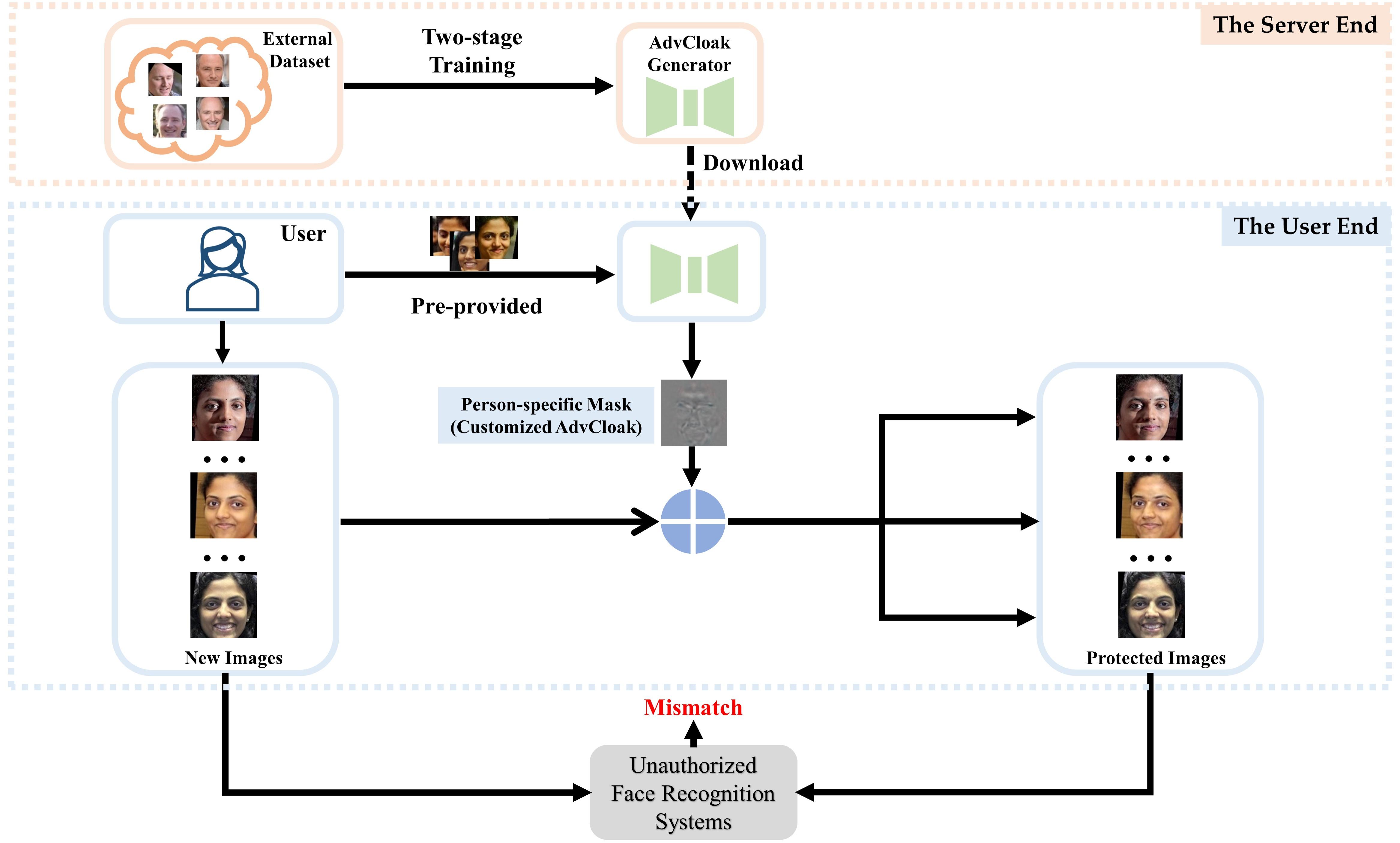}
	\caption{Illustration of person-specific (class-wise) mask customization for privacy protection. Specifically, in the server end, the AdvCloak generator is trained on the external dataset via the proposed two-stage training strategy. In the user end, users are required to prepare several face images and exploit the downloaded AdvCloak generator to synthesize the person-specific masks. This mask can be applied to all the images belonging to an individual, effectively evading identification by face recognition systems.} 
	\label{pipline}
\end{figure}

In recent years, face privacy protection has witnessed notable advancements. Among the various approaches explored, obfuscation techniques~\cite{zhang2014anonymous,letournel2015face,jiang2016face,fan2018image} have received widespread attention. These techniques typically degrade the image quality to render FR systems infeasible, but this degradation often conflicts with users’ intentions of sharing high-quality images on social media~\cite{yang2021towards,zhong2022opom}. Concurrently, another line of research~\cite{chatzikyriakidis2019adversarial,zhang2020adversarial,shan2020fawkes,yang2021towards,cherepanova2021lowkey} advocates for the use of adversarial techniques to subtly alter images, thereby obscuring individual identity information. Nevertheless, generating distinct adversarial masks for huge amounts of face images is an impractical task, hindered by the high computational cost and complexity of crafting instance-specific adversarial examples~\cite{zhong2022opom}.

To tackle these problems, recent research~\cite{zhong2022opom} provides evidence for the existence of person-specific (class-wise) masks that can be applied to all the face images of the individual. This research introduces a method, called OPOM~\cite{zhong2022opom} for generating such masks. OPOM utilizes data priors to model identity-specific feature subspaces and approximates the gradient information by optimizing each adversarial image away from the source identity’s feature
subspace. However, employing multiple gradient queries will limit the transferability towards black-box models, due to the reliance on inappropriate neuron importance measures~\cite{zhang2022improving}. Additionally, adversarial masks optimized based solely on norm constraints often lack visual semantic information, thereby compromising the natural visualization effect~\cite{xiao2018generating, deb2020advfaces}. 

In this paper, we propose a generative model-based framework for face privacy protection, called AdvCloak, which can automatically customize imperceptible privacy masks for each individual. As illustrated in Fig. \ref{pipline}, users can download the trained generative model from the server end and utilize a limited set of their own face images to generate a personalized adversarial cloak. This cloak can then be applied to all new face images to create protected versions at the user end. Specifically, AdvCloak comprises two-stage generators for synthesizing adversarial masks, two-stage discriminators for distinguishing adversarial images from original images, and a surrogate face recognition model to extract facial features. By training on external datasets, AdvCloak can capture the distribution of face images for improved visual naturalness, and learn identity-specific facial salient features for enhanced generalization ability. Moreover, in contrast to OPOM which requires multiple iterations, AdvCloak significantly reduces the average inference time per adversarial mask from 2.81s to 0.05s with a single forward pass. Its simple model structure and low parameter count make it highly practical for deployment at the user end.

AdvCloak achieves person-specific mask generation by examining two primary demands in this challenging task. Firstly, there is a need for image-level naturalness. Due to the inherent variations present in individuals' face images, the data distribution may vary significantly. This diversity poses a challenge in ensuring the naturalness of all the protected face images. Secondly, there is a demand for feature-level generalization ability. The person-specific masks aim to modify the feature distribution across all the protected images. Simultaneously, these privacy masks should be transferable towards different face recognition models. The training mechanism of previous adversarial attack methods based on generative adversarial networks (GANs)~\cite{xiao2018generating,deb2020advfaces} cannot sufficiently address these demands. Therefore, we propose a two-stage training mechanism, consisting of image-specific training and person-specific training. Specifically, in the first stage (image-specific training), we train the networks to adapt to different faces, utilizing a one-image-to-one-mask approach. In the second stage (person-specific training), we introduce a novel training mechanism that learns the salient adversarial features of various face variations within each person for the feature-level generalization. This stage effectively modifies the feature distribution. To further enhance the generalization capability of person-specific masks, we train the networks by optimizing each adversarial image to diverge from the feature subspace associated with the source identity. To fully leverage the training data, we design a distinct feature subspace for each identity by employing various modeling methods such as affine hull, class center, and convex hull~\cite{cevikalp2010face}.

The main contributions of our paper are as follows:
\begin{itemize}
	\item We introduce AdvCloak, a generative model-based approach for automatically synthesizing person-specific (class-wise) masks at the user end. To train AdvCloak, a two-stage training strategy is employed on an external dataset at the server end, encompassing image-specific and person-specific training stages. This two-stage training strategy ensures that the generated masks not only exhibit adaptability to diverse faces but also achieve robust generalization across all the images associated with the source identities at both the image and feature levels.
	\item To enhance the intra-class generalization ability of adversarial masks, we design a unique feature subspace for each source identity and optimize the networks by maximizing the distance between the deep features of adversarial images and their corresponding feature subspaces. 
	\item Both the quantitative and qualitative experiments demonstrate the effectiveness and efficiency of our proposed AdvCloak in face privacy protection towards different black-box face recognition models compared with other methods of generating person-specific masks.
\end{itemize}

\section{RELATED WORKS}
\label{sec:Related}
In this section, we review related works on GANs, adversarial examples and face privacy protection. 

\subsection{Generative Adversarial Networks}
In recent years, GANs have achieved considerable development~\cite{goodfellow2014generative,mirza2014conditional,radford2015unsupervised,gulrajani2017improved}, significantly improving the state-of-the-art in image synthesis. This progress extends to various domains, such as image style transfer~\cite{gatys2016image}, image-to-image translation~\cite{isola2017image},~\cite{zhu2017unpaired} and text-to-image synthesis~\cite{reed2016generative}. Moreover, given the fact that the widely known hard challenges in GANs training, due to the high probability of mode collapse, some valuable works~\cite{isola2017image},~\cite{salimans2016improved} have been proposed to address these issues. In our work, we aim to leverage GANs for synthesizing person-specific masks, imperceptible yet effective, that demonstrate robust generalization across all images of individuals at both the image and feature levels.

\subsection{Adversarial Examples}
Adversarial examples are crafted by intentionally applying minute perturbations to the original instances, leading deep neural networks (DNNs) to be deceived with a high level of confidence. Existing approaches for generating adversarial examples can be categorized into two types: image-specific adversarial examples and universal adversarial examples (image-agnostic).

First, one of mainstream approaches involves crafting image-specific adversarial examples~\cite{szegedy2013intriguing,goodfellow2014explaining,kurakin2018adversarial,dong2018boosting,carlini2017towards,xiao2018generating}. Szegedy \textit{et al.}~\cite{szegedy2013intriguing} firstly proposed the concept of adversarial examples and a box-constrained method, L-BFGS, to generate them. Goodfellow \textit{et al.}~\cite{goodfellow2014explaining} introduced a fast gradient sign method (FGSM) by performing a single gradient step, which is easier and faster to compute. Based on FGSM, Kurakin \textit{et al.}~\cite{kurakin2018adversarial} further proposed the iterative method named as I-FGSM. Furthermore, Dong \textit{et al.}~\cite{dong2018boosting} proposed an MI-FGSM method by adding the momentum term to it. From another point of view, Carlini \textit{et al.}~\cite{carlini2017towards} proposed an optimization-based method by using the Carlini-Wagner (CW) loss function. To generate more natural adversarial examples, Xiao \textit{et al.}~\cite{xiao2018generating} proposed to train the conditional adversarial networks, AdvGAN, to directly synthesize them.

Since image-specific adversarial perturbations vary for different images in a dataset, there exists a universal perturbation~\cite{moosavi2017universal,mopuri2018generalizable,poursaeed2018generative,hayes2018learning,xu2020generating,gupta2019method,zhang2020cd, liu2023enhancing} for all the images of the dataset. Moosavi-Dezfooli \textit{et al.}~\cite{moosavi2017universal} demonstrated the existence of a universal perturbation (UAP) which could be added to all the images of the database to fool the classifiers with only one elaborate design. Mopuri \textit{et al.}~\cite{mopuri2018generalizable} proposed a novel data-free method (GD-UAP) for generating universal perturbations. Poursaeed et at.~\cite{poursaeed2018generative} and Hayes et at.~\cite{hayes2018learning} both introduced generative models for crafting universal adversarial examples. Xu \textit{et al.}~\cite{xu2020generating} proposed a framework that exploited a residual network (ResNet) to create universal adversarial perturbations, named as UAP-RN. In addition to the universal adversarial examples, Gupta \textit{et al.}~\cite{gupta2019method} proposed a method for crafting class-wise adversarial perturbations which didn’t require any training data and hyper-parameters via the linearity of the decision boundaries of DNNs. Zhang \textit{et al.}~\cite{zhang2020cd} firstly proposed a new type of class discriminative universal adversarial example (CD-UAP), which could fool a network to misclassify the selected classes without affecting other classes. Liu \textit{et al.}~\cite{liu2023enhancing} proposes to aggregate multiple small-batch gradients to alleviate the gradient vanishing. Similar to class-wise methods, we aim to customize a person-specific (class-wise) mask for each user to hide the identity information of all his or her images. 

\subsection{Face Privacy Protection}
The wide spread of face recognition technology is increasingly perceived as a substantial risk to individual privacy and security. Therefore, researchers have delved into studies on face privacy protection.

Among the existing techniques, obfuscation-based methods have been widely explored. A recent work~\cite{meden2021privacy} divided the obfuscation methods into three subsets: masking~\cite{zhang2014anonymous}, filtering~\cite{letournel2015face} and image transformation~\cite{jiang2016face},~\cite{fan2018image}. Techniques from this group are generally on the basis of computationally simple privacy mechanisms, making them widely available. However, previous work~\cite{mcpherson2016defeating} proved that the images protected by the obfuscation techniques might be identified by current face recognition systems. In addition, obfuscated images usually hide sensitive information at the expense of degrading the image quality, which is contrary to the original intentions of users' sharing on social media.

Considering the unnaturalness of obfuscation images, some researchers applied adversarial example techniques to face privacy protection~\cite{chatzikyriakidis2019adversarial,zhang2020adversarial,shan2020fawkes,yang2021towards,cherepanova2021lowkey,zhong2022opom}. Chatzikyriakidis \textit{et al.}~\cite{chatzikyriakidis2019adversarial} introduced a novel adversarial attack method P-FGVM which both protected privacy and kept facial image quality. Zhang \textit{et al.}~\cite{zhang2020adversarial} designed an adversarial privacy-preserving filter (APF) that protected users’ privacy from the malicious identification by face recognition systems. In addition, APF was an end-cloud collaborative framework to make sure that only users’ own devices could access the original images. Fawkes~\cite{shan2020fawkes} fooled face recognition systems at the inference stage by injecting tiny and imperceptible perturbations into the training data. Yang \textit{et al.}~\cite{yang2021towards} proposed a targeted identity-protection iterative method (TIP-IM) which took targeted protection and unknown gallery set into consideration. Cherepanova \textit{et al.}~\cite{cherepanova2021lowkey} proposed to add adversarial perturbations to gallery faces so that probe images cannot be correctly identified.

At present, AdvFaces~\cite{deb2020advfaces} effectively ensures the image naturalness by utilizing GANs to synthesize visually realistic adversarial face images. However, AdvFaces and the above methods of privacy protection~\cite{chatzikyriakidis2019adversarial,zhang2020adversarial,shan2020fawkes,yang2021towards,cherepanova2021lowkey} are image-specific methods that are impractical to generate corresponding adversarial perturbations for vast amount of face images. To address these problems, Zhong \textit{et al.}~\cite{zhong2022opom} firstly demonstrated the existence of person-specific masks (class-wise) to protect various face images of one person and proposed a gradient-based method, OPOM, to generate this type of adversarial mask. However, the OPOM method employs multiple gradient queries to alter every pixel in the image, which lacks the visual semantic information and limits the transferability towards black models owing to the dependence on inappropriate neuron importance measures. Moreover, OPOM requires unreasonable time to customize this type of mask and inevitably needs to upload images to the server which has the risk of image leakage. Different from previous works, in our work, we explore employing the generative models to automatically customize the adversarial cloaks (person-specific masks) which take into account both the image-level naturalness and feature-level generalization ability. Furthermore, the trained generative model can be downloaded by the user to the local end for use without requiring unreasonable time to query the server and perform gradient backpropagation like in OPOM~\cite{zhong2022opom}. 

\begin{figure*}[!ht]
	\centering
	\includegraphics[scale=0.33]{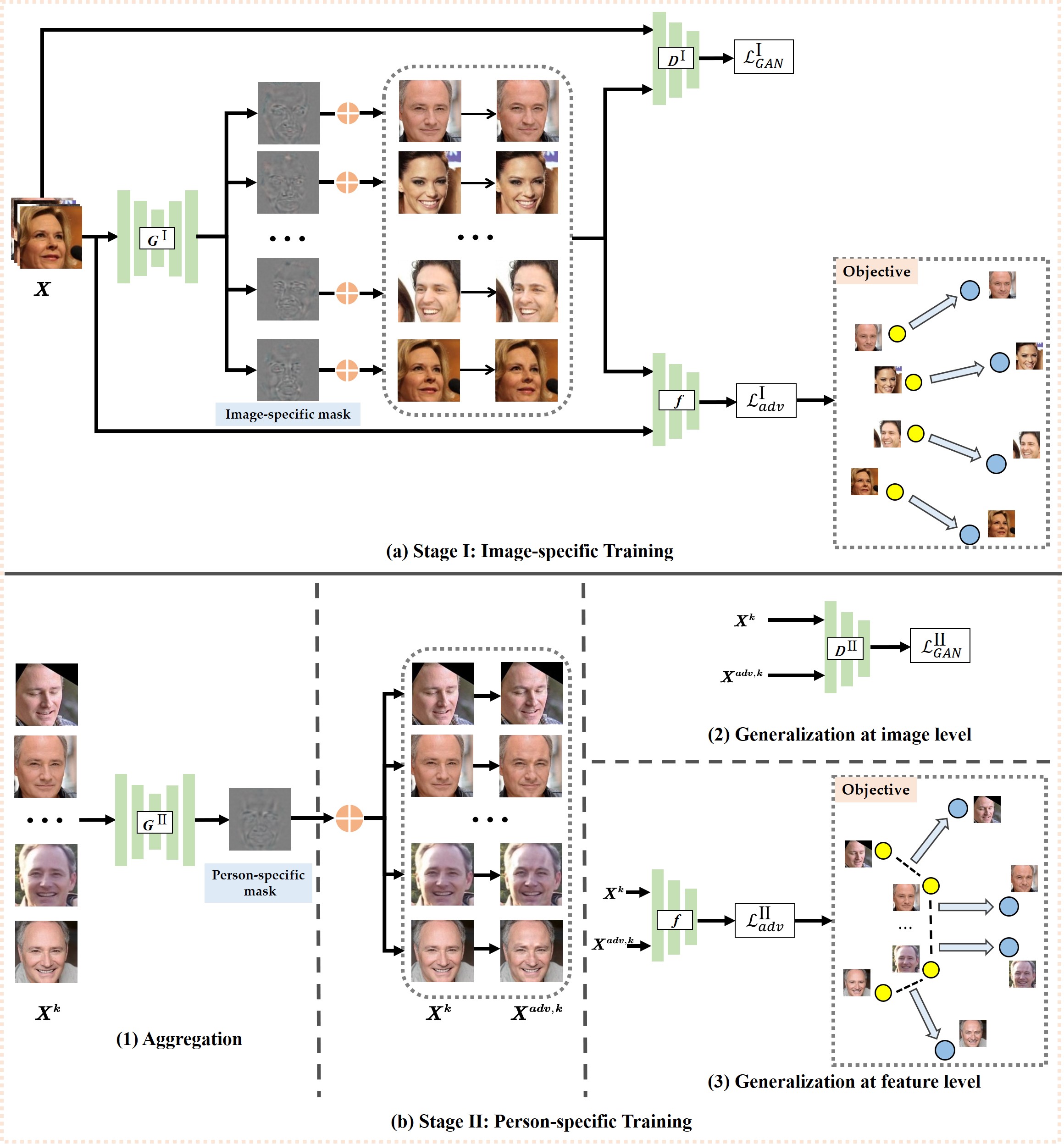}
	\caption{Illustration of the overall framework of AdvCloak. (a) The top of the figure illustrates the image-specific training stage of AdvCloak, enabling adaptation to different faces. (b) The bottom of the figure illustrates the person-specific training stage of AdvCloak to make the generated person-specific (class-wise) masks generalize well to all the images of the source identity. To better improve the intra-class generalization ability of the adversarial masks, the networks are optimized by maximizing the distance between the deep features of adversarial images and the feature subspace of the source identity.}
	\vspace{-4mm}
	\label{AdvCloak}
\end{figure*}

\section{Automatically Customized Adversarial Cloak}
\label{sec:Method}
In this section,  we first introduce the problem formulation of the person-specific adversarial mask, then we describe our AdvCloak method and the potential comparable methods.

\subsection{Problem Formulation}
Let $X^k=\left\{ X_{1}^{k},X_{2}^{k},...,X_{i}^{k},... \right\}$ be a face image set belonging to identity $k$, we aim to craft a person-specific adversarial mask $\Delta ^k$, which can be added to all the face images sampled from $X^k$ to evade the recognition by deep face recognition models $f\left( \cdot \right) $. The objective of the above adversarial mask can be formulated as:
\begin{equation}
	\label{Opti}
	\begin{aligned}
		\ell _d\left( f\left( X_{i}^{k}+\Delta ^k \right) ,f\left( X_{i}^{k} \right) \right) >t, \quad &||\Delta ^k||_p\leqslant \epsilon,  \\
		for\,\,most\,\,X_{i}^{k}\in X^k&,
	\end{aligned}
\end{equation}
where $f\left( X_{i}^{k} \right) \in \mathbb{R} ^d$ is the feature vector of $i^{th}$ image in $X^k$, $\ell _d\left( x_1,x_2 \right)$ measures the distance between $x_1$ and $x_2$, $t$ is the distance threshold to decide whether a pair of face images belongs to the same identity, $\epsilon$ controls the magnitude of the adversarial mask.

To solve the above objective, we propose to exploit generative models to capture the data distribution for visual naturalness and learn salient features for enhancing generalization ability. For this purpose, we propose AdvCloak, which is trained on an external dataset through a two-stage training strategy. The proposed two-stage training strategy involves image-specific training and person-specific training to sequentially optimize the generative networks. In the first stage, AdvCloak optimizes the networks for adapting to different persons by implementing the image-specific training mechanism. In the second stage, the person-specific training mechanism is designed to facilitate the generalization of the person-specific mask across diverse images for each individual. Additionally, we incorporate the optimization objective of AdvCloak with modeling methods of the feature subspace for amplifying the intra-class generalization capability of adversarial masks. Details of the proposed AdvCloak are provided in the following Section \ref{details}.

\subsection{The Proposed AdvCloak}
\label{details}
Figure \ref{AdvCloak} illustrates the details of the proposed AdvCloak. The networks consist of two generators $G$, two discriminators $D$, and a deep face recognition model $f$. In different stages, we separately term generator and discriminator as
$\left\{ G^{\mathrm{\uppercase\expandafter{\romannumeral1}}}, G^{\mathrm{\uppercase\expandafter{\romannumeral2}}} \right\}$ and $\left\{ D^{\mathrm{\uppercase\expandafter{\romannumeral1}}}, D^{\mathrm{\uppercase\expandafter{\romannumeral2}}} \right\}$.

\begin{itemize}
	\item \textbf{Generator $\boldsymbol{G^{\mathrm{\uppercase\expandafter{\romannumeral1}}}}$} is designed to utilize original face images to synthesize instance-specific adversarial perturbations by implementing the image-specific training (one image one mask) as shown in Figure \ref{AdvCloak}(a). 
	\item \textbf{Generator $\boldsymbol{G^{\mathrm{\uppercase\expandafter{\romannumeral2}}}}$} aims to aggregate all the perturbations belonging to the same person to obtain the person-specific mask for each person by employing a person-specific training mechanism (one person one mask) as shown in Figure \ref{AdvCloak}(b)-(1). This way, by calculating the adversarial loss between the generated adversarial face images and the ground truth, the generator network can learn the facial salient regions that can be perturbed minimally to hide the individual identity information.
	\item \textbf{Discriminator $\boldsymbol{D^{\mathrm{\uppercase\expandafter{\romannumeral1}}}}$} competes against $G^{\mathrm{\uppercase\expandafter{\romannumeral1}}}$ to distinguish original images and adversarial images, while $G^{\mathrm{\uppercase\expandafter{\romannumeral1}}}$ aims to fool $D^{\mathrm{\uppercase\expandafter{\romannumeral1}}}$. The goal of $D^{\mathrm{\uppercase\expandafter{\romannumeral1}}}$ is to ensure that the adversarial face images look real. This way, it enables the generator network to have an ability of deep semantic understanding, thereby avoiding destroying the naturalness of adversarial face images. In stage \uppercase\expandafter{\romannumeral1}, they are optimized by the image-wise GAN loss $\mathcal{L} _{GAN}^{\mathrm{\uppercase\expandafter{\romannumeral1}}}$ in Figure \ref{AdvCloak}(a).
	\item \textbf{Discriminator $\boldsymbol{D^{\mathrm{\uppercase\expandafter{\romannumeral2}}}}$} competes against $G^{\mathrm{\uppercase\expandafter{\romannumeral2}}}$ to distinguish original images and adversarial images, while $G^{\mathrm{\uppercase\expandafter{\romannumeral2}}}$ aims to fool $D^{\mathrm{\uppercase\expandafter{\romannumeral2}}}$. $D^{\mathrm{\uppercase\expandafter{\romannumeral2}}}$ aims to force the generated person-specific masks to generalize to all the images of the source identity at the image level. In stage \uppercase\expandafter{\romannumeral2}, they are optimized by the class-wise GAN loss $\mathcal{L} _{GAN}^{\mathrm{\uppercase\expandafter{\romannumeral2}}}$ in Figure \ref{AdvCloak}(b)-(2).
	\item \textbf{Face Recognition Model $\boldsymbol{f}$} is a pre-trained model used as a surrogate model to mimic unknown face recognition systems. With the elaborate image-wise adversarial loss $\mathcal{L} _{adv}^{\mathrm{\uppercase\expandafter{\romannumeral1}}}$ in Figure \ref{AdvCloak}(a) and class-wise adversarial loss $\mathcal{L} _{adv}^{\mathrm{\uppercase\expandafter{\romannumeral2}}}$ in Figure \ref{AdvCloak}(b)-(3), $f$ is used to extract deep features of the adversarial images and the original images, and provide adversarial information to optimize the networks for concealing the identity information under the perturbations.
\end{itemize}

\subsubsection{Image-specific and Person-specific Training}
\textbf{Image-specific Training.} In stage \uppercase\expandafter{\romannumeral1}, the optimization goal of the networks is to synthesize the corresponding adversarial perturbations to adapt to different faces. In this way, we implement an image-specific training mechanism. Specifically, let $X=\left\{ X_1,X_2,...,X_i,... \right\} $ be the random face image set. The corresponding perturbation generated from $X_i$ is defined as $G^{\mathrm{\uppercase\expandafter{\romannumeral1}}}\left( X_i \right)$ and the adversarial face image $X_{i}^{adv}$ is defined as $X_i+G^{\mathrm{\uppercase\expandafter{\romannumeral1}}}\left( X_i \right)$. Through such training, the $G^{\mathrm{\uppercase\expandafter{\romannumeral1}}}$ can map an original image $X_i$ to a tiny perturbation with high semantic understanding. Such perturbation is invisible when applied to the corresponding face image but can significantly change the distribution of the deep feature. However, the generated perturbations exhibit high correlations with the original images, limiting their ability to generalize effectively to new face images of the same identity. Consequently, we proceed to optimize the network for the acquisition of class-specific feature representations.

\textbf{Person-specific Training.} To enhance the generalization of the privacy masks across all images of the source identity, we propose a novel training mechanism, termed person-specific training. This approach, built upon image-specific training, simultaneously learns aggregation and generalization processes, providing additional optimization to the networks. Specifically, the generator $G^{\mathrm{\uppercase\expandafter{\romannumeral2}}}$ takes a set of face images of the identity $k$, $X^k$, as input, and the person-specific mask is obtained by aggregating the perturbations generated by all the images of $X^k$:
\begin{equation}
	\label{generation}
	\Delta ^k=\frac{1}{n}\sum_{i=1}^n{G^{\mathrm{\uppercase\expandafter{\romannumeral2}}}\left( X_{i}^{k} \right)},
\end{equation}
where $X_{i}^{k}$ is the $i^{th}$ image in $X^k$ and $n$ is the number of images in $X^k$. For each person, the adversarial face images are formed by adding the adversarial mask to all the original images:
\begin{equation}
	X_{i}^{adv,k}=X_{i}^{k}+\Delta ^k, \quad for\,\,i=1,2,...,n,
\end{equation}
where $	X_{i}^{adv,k}$ is the $i^{th}$ image in adversarial face image set $X^{adv,k}$. Then we feed the original image set $X^k$ and adversarial image set $X^{adv,k}$ to both $D^{\mathrm{\uppercase\expandafter{\romannumeral2}}}$ and $f$ to make the distribution of the source and target domain similar at the image level but far from each other at the feature level. This approach captures universal adversarial information, effectively modifying the feature distributions and facilitating feature-level generalization.

\subsubsection{Modeling of Feature Subspace}
\label{feature}
Furthermore, the person-specific training stage brings an additional advantage: using $n$ training face images within the set $X^k$ corresponding to identity $k,$ we can construct a feature subspace for each individual, thereby increasing the diversity of deep features.

When generating person-specific masks, an intuitive idea is that the more images provided by each user, the higher the privacy protection performance achieved by the generated adversarial mask. Nevertheless, practical scenarios often pose challenges in acquiring a substantial number of face images. Therefore, it is more feasible to consider customizing this type of mask using a limited number of images. In this way, Constructing the feature subspace can help better describe the feature distribution of the identity. To enhance the intra-class generalization ability of person-specific masks, we attempt to reformulate the optimization problem in Eq. \ref{Opti} by maximizing the distance between deep features of each adversarial face image and the feature subspace associated with the identity,
\begin{equation}
	\label{subspace}
	\begin{aligned}
		\ell _d\left( f\left( X_{i}^{k}+\Delta ^k \right) ,H_{X^k} \right) >t, \quad &||\Delta ^k||_p\leqslant \epsilon,  \\
		for\,\,most\,\,X_{i}^{k}\in X^k,&
	\end{aligned}
\end{equation}
where $H_{X^k}$ represents the feature subspace of identity $k$. Next, we will introduce some modeling methods of feature subspace.

\textbf{Affine Hull~\cite{cevikalp2010face}.} Firstly, for a specific identity denoted as $k$, the feature subspace can be approximated by the affine hull of deep features of his or her images. This approximation is formulated as:
\begin{equation}
	\label{aff}
	\begin{aligned}
	H_{k}^{Aff}=\left\{ x=\sum_{i=1}^n{\alpha _{i}^{k}f( X_{i}^{k} ) |\sum_{i=1}^n{\alpha _{i}^{k}=1}} \right\}, 
	\end{aligned}
\end{equation}		
where $f\left( X_{i}^{k} \right) \in \mathbb{R} ^d$ is the feature vector of $i^{th}$ image in $X^k$, $\alpha _{i}^{k}\in R$ are the weighting factors to describe the identity $k$.

\textbf{Class Center.} The feature subspace modeled by the affine hull may be excessively large, resulting in some points of the hull not being features of the identity $k$, and thus has a bad effect on the generation of the person-specific masks. A naive solution is that the feature subspace can be described as the mean feature, and it is defined as:
\begin{equation}
	\label{cen}
	H_{k}^{Cen}=\frac{1}{n}\sum_{i=1}^n{f\left( X_{i}^{k} \right)}.
\end{equation}
	
\textbf{Convex Hull~\cite{cevikalp2010face}.} Another solution is that we make upper and lower bound constraints on the weighting factor $\alpha ^k$ in the modeling of feature subspace. An affine hull is a special case when the lower bound is $-\infty$, upper bound is $+\infty$. Adding bound constraints can effectively control the regions of the feature subspace. Furthermore, there exists a convex hull (the smallest convex set) of modeling feature subspace when the lower bound is $0$, the upper bound is $1$. The convex hull has been confirmed to be much tighter than the affine approximation~\cite{cevikalp2010face} to solve the problem that the affine hull is too loose. In this way, the convex hull is formulated as:
\begin{equation}
	\label{con}
	H_{k}^{Con}=\left\{ x=\sum_{i=1}^n{\alpha _{i}^{k}f( X_{i}^{k} ) |\sum_{i=1}^n{\alpha _{i}^{k}=1,0\leqslant \alpha _{i}^{k}\leqslant 1}} \right\}.
\end{equation}

%-------------------------------------------------------------------------
\subsubsection{Objective Functions}
The total loss of $\mathcal{L}$ consists of GAN loss $\mathcal{L} _{GAN}$, norm constraint loss $\mathcal{L} _{norm}$, adversarial loss $\mathcal{L} _{adv}$. $\mathcal{L}$ is defined as:
\begin{equation}
	\mathcal{L} =\mathcal{L} _{GAN}+\lambda _{norm}\mathcal{L} _{norm}+\lambda _{adv}\mathcal{L} _{adv},
\end{equation}
where $\lambda _{norm}$ and $\lambda _{adv}$ are hyper-parameters that represent relative weights for the loss function. Therefore, we optimize $G$ and $D$ in a min-max manner with the following objectives:
\begin{equation}
	\underset{G}{\min}\,\,\underset{D}{\max}\,\,\mathcal{L} _{GAN}+\lambda _{norm}\mathcal{L} _{norm}+\lambda _{adv}\mathcal{L} _{adv}.
\end{equation}
The following are the details of each loss function.

\textbf{GAN Loss.} In stage \uppercase\expandafter{\romannumeral1}, we employ the image-wise GAN loss $\mathcal{L} _{GAN}^{\mathrm{\uppercase\expandafter{\romannumeral1}}}$ to encourage the generator $G^{\mathrm{\uppercase\expandafter{\romannumeral1}}}$ to learn the salient regions for improving the adaptation of the generated perturbations to various individual faces. Concurrently, the discriminator $D^{\mathrm{\uppercase\expandafter{\romannumeral1}}}$ is designed to classify whether the input images belong to the original set or are synthesized adversarial images. These components can be mathematically expressed as follows:
\begin{equation}
		\mathcal{L} _{GAN}^{\mathrm{\uppercase\expandafter{\romannumeral1}}}=\mathbb{E} _{X}\left[ \log D\left( X \right) \right] + \mathbb{E} _{X}\left[ \log \left( 1-D\left( X+G\left( X \right) \right) \right) \right]. 
\end{equation}
In stage \uppercase\expandafter{\romannumeral2}, for enhancing the generalization ability of the generated person-specific masks to all the images of the same person at the image level, we employ the class-wise GAN loss $\mathcal{L} _{GAN}^{\mathrm{\uppercase\expandafter{\romannumeral2}}}$ to optimize $G^{\mathrm{\uppercase\expandafter{\romannumeral2}}}$ and $D^{\mathrm{\uppercase\expandafter{\romannumeral2}}}$. They can be expressed as:
\begin{equation}
		\mathcal{L} _{GAN}^{\mathrm{\uppercase\expandafter{\romannumeral2}}}=\mathbb{E} _k\left[ \frac{1}{n}\sum_{i=1}^n{\log D\left( X_{i}^{k} \right)} \right] + \mathbb{E} _k\left[ \frac{1}{n}\sum_{i=1}^n{\log \left( 1-D\left( X_{i}^{k}+\Delta _k \right) \right)} \right]. 
\end{equation}

\textbf{Norm Constraint Loss.} 
$\mathcal{L} _{GAN}$ focuses only on aligning the adversarial images with the data distribution of original images, neglecting consideration for the perturbation magnitude of the adversarial mask. To address this, we introduce $\mathcal{L} _{norm}^{\mathrm{\uppercase\expandafter{\romannumeral1}}}$ to constrain the $L_2$ norm of the image-wise perturbations and $\mathcal{L} _{norm}^{\mathrm{\uppercase\expandafter{\romannumeral2}}}$ to constrain the $L_2$ norm of the class-wise perturbations. They can be formulated as:
\begin{equation}
	\mathcal{L} _{norm}^{\mathrm{\uppercase\expandafter{\romannumeral1}}}=\mathbb{E} _{X}\left[ \max \left( \epsilon ,||G\left( X \right) ||_2 \right) \right],
\end{equation}
\begin{equation}
	\mathcal{L} _{norm}^{\mathrm{\uppercase\expandafter{\romannumeral2}}}=\mathbb{E} _k\left[ \max \left( \epsilon ,||\Delta ^k||_2 \right) \right],
\end{equation}
where $\epsilon$ represents a hyperparameter controlling the perturbation magnitude of the adversarial masks.

\textbf{Adversarial Loss.} We calculate the feature-level distance between the original images and the adversarial images utilizing the face recognition model $f$. Then, we employ the image-wise adversarial loss $\mathcal{L} _{adv}^{\mathrm{\uppercase\expandafter{\romannumeral1}}}$ to maximize the feature distance between the adversarial images and the original images. The formulation for the image-wise adversarial loss is as follows:
\begin{equation}
	\mathcal{L} _{adv}^{\mathrm{\uppercase\expandafter{\romannumeral1}}}=\mathbb{E} _{X}\left[ \ell _d\left( f\left( X+G\left( X \right) \right) ,f\left( X \right) \right) \right].
\end{equation}
Also, to make the generated person-specific masks better generalize to all the images of the same person at the feature level, we introduce the class-wise adversarial loss, denoted as $\mathcal{L} _{adv}^{\mathrm{\uppercase\expandafter{\romannumeral2}}}$, into the network optimization process. The formulation for the class-wise adversarial loss is as follows:
\begin{equation}
	\mathcal{L} _{adv}^{\mathrm{\uppercase\expandafter{\romannumeral2}}}=\mathbb{E} _k\left[ \frac{1}{n}\sum_{i=1}^n{\ell _d\left( f\left( X_{i}^{k}+\Delta ^k \right) ,f\left( X_{i}^{k} \right) \right)} \right].  
\end{equation}

As previously discussed in Section \ref{feature}, the modeling of the feature subspace proves to be more effective in characterizing identity. Therefore, in stage $\mathrm{\uppercase\expandafter{\romannumeral2}}$, combined with Eq. (\ref{subspace}), the optimization objective of the class-wise adversarial loss $\mathcal{L} _{adv}^{H_{X^k}}$ aims to force each adversarial image in the direction away from the feature subspace of the source identity instead of the corresponding original images. The $\mathcal{L} _{adv}^{H_{X^k}}$ can be formulated as:
\begin{equation}
	\mathcal{L} _{adv}^{H_{X^k}}=\mathbb{E} _k\left[ \frac{1}{n}\sum_{i=1}^n{\ell _d\left( f\left( X_{i}^{k}+\Delta ^k \right) ,H_{X^k} \right)} \right],  
\end{equation}
where $H_{X^k}$ represents the feature subspace of the identity $k$. We incorporate various modeling methods for the feature subspace including affine hull, class center, and convex hull. These methods are implemented in the $\mathcal{L} _{adv}^{H_{X^k}}$, resulting in three distinct forms: $\mathcal{L} _{adv}^{Aff}$, $\mathcal{L} _{adv}^{Cen}$, and $\mathcal{L} _{adv}^{Con}$. They can be described as follows:
\begin{equation}
	\begin{aligned}
		\mathcal{L} _{adv}^{Aff}=\mathbb{E} _k\left[ \frac{1}{n}\underset{\alpha ^k}{\min}\sum_{i=1}^n{\ell _d\left( f\left( X_{i}^{k}+\Delta ^k \right) ,H_{k}^{Aff} \right)} \right],\\
		H_{k}^{aff}=\left\{ x=\sum_{i=1}^n{\alpha _{i}^{k}f( X_{i}^{k} ) |\sum_{i=1}^n{\alpha _{i}^{k}=1}} \right\}, 
	\end{aligned}
\end{equation}
\begin{equation}
	\mathcal{L} _{adv}^{Cen}=\mathbb{E} _k\left[ \frac{1}{n}\sum_{i=1}^n{\ell _d\left( f\left( X_{i}^{k}+\Delta ^k \right) ,\frac{1}{n}\sum_{i=1}^n{f\left( X_{i}^{k} \right)} \right)} \right],
\end{equation}
\begin{equation}
	\begin{aligned}
		\mathcal{L} _{adv}^{Con}=\mathbb{E} _k\left[ \frac{1}{n}\underset{\alpha ^k}{\min}\sum_{i=1}^n{\ell _d\left( f\left( X_{i}^{k}+\Delta ^k \right) ,H_{k}^{Con} \right)} \right],\\
		H_{k}^{Con}=\left\{ x=\sum_{i=1}^n{\alpha _{i}^{k}f( X_{i}^{k} ) |\sum_{i=1}^n{\alpha _{i}^{k}=1}},0\leqslant \alpha _{i}^{k}\leqslant 1 \right\}. 
	\end{aligned}
\end{equation}
The distance between the features of adversarial images and feature subspace can be efficiently solved with the convex optimization toolbox CVX~\cite{diamond2016cvxpy},~\cite{agrawal2018rewriting}. Finally, by integrating approximation methods for the feature subspace, including affine hull, class center, and convex hull, three distinct variants of AdvCloak can be named as AdvCloak-AffineHull, AdvCloak-ClassCenter, and AdvCloak-ConvexHull respectively.

\subsubsection{Generation of Person-specific Masks}
Following the training of AdvCloak on the server, users have the option to download and store the generator model locally. Subsequently, each user can utilize the trained model to generate a person-specific mask based on Eq. (\ref{generation}), leveraging a limited set of pre-provided face images.

\subsection{Comparison methods}
In addition to AdvCloak, we investigate alternative methods for generating universal adversarial perturbations.

\subsubsection{GAP}
Generative Adversarial Perturbations (GAP)~\cite{poursaeed2018generative} use generative models to generate universal adversarial perturbations which can be applied to all the images of the dataset to misclassify models. 

\subsubsection{AdvFaces+}
AdvFaces~\cite{deb2020advfaces} is a GAN-based method to synthesize realistic adversarial face images for fooling the face recognition systems. However, the perturbation generated by AdvFaces is a kind of image-specific perturbation. To generate a person-specific mask, one straightforward thought is to average perturbations generated from all the inference images belonging to the same person to obtain the person-specific masks, named as AdvFaces+.

\subsubsection{FI-UAP}
FI-UAP~\cite{zhong2022opom} is a gradient-based method to generate a person-specific mask. Similar to the process in UAP~\cite{moosavi2017universal}, which iteratively aggregates the minimal perturbations generated by DeepFool~\cite{moosavi2016deepfool} to craft universal adversarial perturbations, FI-UAP aggregate perturbations generated by feature iterative attack method (FIM)~\cite{zhong2020towards},~\cite{zhong2019adversarial}:
\begin{equation}
	\begin{split}
		&\Delta _{0}^{k}=0,\\
		&g_t=\frac{1}{n}\sum_{i=1}^n{\nabla _x\ell _d\left( f\left( X_{i}^{k}+\Delta _{t}^{k} \right) ,f\left( X_{i}^{k} \right) \right)},\\
		&\Delta _{t+1}^{k}=Clip_{\varepsilon}\left( \Delta _{t}^{k}+\alpha sign\left( g_t \right) \right).  
	\end{split}
\end{equation}
Also, we combined the FI-UAP with the modeling methods of feature subspace in Eq. (\ref{aff}), Eq. (\ref{cen}), and Eq. (\ref{con}), named as OPOM-AffineHull, OPOM-ClassCenter, OPOM-ConvexHull for comparison~\cite{zhong2022opom}.

\section{EXPERIMENTS}
\label{experiments}
\subsection{Experimental Settings}
\subsubsection{Databases} AdvCloak is first trained on an external dataset, CASIA-WebFace~\cite{yi2014learning}. Then we evaluate our methods on two datasets, \textit{i.e.}, Privacy-Commons dataset, and Privacy-Celebrities dataset.

The CASIA-WebFace~\cite{yi2014learning} database consists of 494,414 face images belonging to 10,575 different identities. In the image-specific training stage, we choose 50,000 face images from the CASIA-WebFace database and randomly shuffle the order. In the person-specific training stage, we select 3,000 persons in the CASIA-WebFace database, each with 16 images, for a total of 48,000 original images.

The MegaFace challenge2~\cite{nech2017level} is a large publicly available database to evaluate the performance of face models at the million-scale distractors. We select 500 people in the MegaFace challenge2 database, each with a dataset of 15 clean images (without label noise) to combine the Privacy-Commons dataset. For each selected identity, the images in the dataset are split into two subsets, one for ten inference images and the other for five test images. We can use the inference images as data prior to generate the person-specific (class-wise) mask like in OPOM~\cite{zhong2022opom}; while the other subset is used as unknown images of the source identity to evaluate protection performance. In addition, we select 10,000 images of other identities in the training dataset as distractors to compose the gallery set. 

Likewise, we select 500 people from the one-million celebrity list, each with a dataset of 20 clean images (without label noise), without being repeated in MS-Celeb-1M~\cite{guo2016ms} and LFW database~\cite{huang2008labeled}, which are combined into the Privacy-Celebrities dataset. For each selected identity, the images in the dataset are split into two subsets, one for 10 inference images and the other for 10 test images. We can use the inference images to customize the person-specific mask; while the other subset for testing is used as unknown images of the source identity to evaluate protection performance. We use a celebrity database, 13,233 images of LFW database~\cite{huang2008labeled} as distractors to compose the gallery set.

\subsubsection{Evaluation Metrics} In our paper, we focus on a general scenario in the open-set face identification that we report 1:N recognition performance to evaluate the protection success rate. Specifically, 1:N recognition is achieved by identifying images from the gallery set with the same person as the probe image. For each person, we take turns putting each image into the gallery set and the rest of the images are used as probe images. The algorithm will rank the feature distances between the probe image and the images in the gallery set. Unlike the previous works adding perturbations to training data~\cite{shan2020fawkes} or gallery set~\cite{cherepanova2021lowkey}, we add the privacy masks to the probe images. Since we assume that the user's original images uploaded to the social platform have been obtained by unauthorized third organizations or individuals and added to the gallery for malicious identification. We report the Top-1 and Top-5 protection success rates (100\%-Top1 or Top5 accuracy) to evaluate the privacy protection performance. Taking the Privacy-Commons dataset as an example, we construct $500\times 5\times 4=10,000$ test pairs ($500\times 10\times 9=45,000$ for the Privacy-Celebrities dataset) and calculate the ratio of the identification success pairs with Top-1 and Top-5 images having the same identity as probe image to the total test pairs.

\subsubsection{Face Recognition Models}
In our paper, we craft person-specific masks for each identity against different source models which are supervised by ArcFace~\cite{deng2019arcface}, CosFace~\cite{wang2018cosface}, and softmax loss. In the actual privacy protection scenario, our goal is to evade recognition by the black-box face recognition systems. Therefore, we use six black-box models as target models that differ in loss functions and network architectures respectively. The former contains ArcFace~\cite{deng2019arcface}, CosFace~\cite{wang2018cosface} and SFace~\cite{zhong2021sface}, the latter includes MobileNet~\cite{howard2017mobilenets}, SENet~\cite{hu2018squeeze} and Inception-ResNet (IncRes)~\cite{szegedy2017inception}.

\subsubsection{Implementation Details}
We illustrate the training process of AdvCloak in Figure \ref{AdvCloak}. The hyper-parameters $\epsilon$ for controlling the perturbation is set to be 3, which aligns with the parameter configurations employed in AdvFaces~\cite{deb2020advfaces}. In the image-specific training stage, the $\lambda _{norm}$ and $\lambda _{adv}$ are set to be 1 and 10. And in the person-specific training stage, the $\lambda _{norm}$ and $\lambda _{adv}$ are set to be 1 and 15. We use ADAM optimizers in Pytorch with the exponential decay rates as $\left( \beta _1,\beta _2 \right) =\left( 0.5,0.9 \right)$. The learning rate of the image-specific training stage is set to be 0.0001 and $2\times 10^{-8}$ is set in the person-specific training stage for finetuning the network.

For other compared methods, some details are as follows. For FI-UAP and OPOM~\cite{zhong2022opom} methods, we follow the parameter settings in the original paper which set the maximum deviation of perturbations $\varepsilon$ to 8 under the $L_{\infty}$ constraint and the number of iterations to 16. For AdvFaces+, the hyperparameter settings follow the obfuscation attack in the AdvFaces~\cite{deb2020advfaces}.

Considering that the perturbation magnitudes of the adversarial masks generated by different methods differ, we constrain the $l_2$ norm of all the masks to 1,200 to establish a valid comparison.

\begin{table*}[]
\setlength\tabcolsep{2.5pt}
	\renewcommand{\arraystretch}{1.40}
	\caption{\centering Top-1 and Top-5 protection success rates (\%) are reported against black-box models under the 1:N identification setting on the Privacy-Commons dataset. The AdvCloak-based methods are our proposed methods.}
	\label{ten}
        \vspace{-2mm}
	\resizebox{\linewidth}{!}{
		\begin{tabular}{c|c|cccccccccccc|cc}
			\toprule
			\multirow{3}{*}{\textbf{\begin{tabular}[c]{@{}c@{}}Source\\ Model\end{tabular}}} & \multirow{3}{*}{\textbf{Method}}     & \multicolumn{12}{c|}{\textbf{Target Model}}                                                                                                                                                                                                                                                                                       & \multicolumn{2}{c}{\multirow{2}{*}{\textbf{Average}}} \\ \cline{3-14}
			&                                      & \multicolumn{2}{c|}{\textbf{ArcFace}}                & \multicolumn{2}{c|}{\textbf{CosFace}}                & \multicolumn{2}{c|}{\textbf{SFace}}                  & \multicolumn{2}{c|}{\textbf{MobileNet}}              & \multicolumn{2}{c|}{\textbf{SENet}}                  & \multicolumn{2}{c|}{\textbf{IncRes}} & \multicolumn{2}{c}{}                                  \\ \cline{15-16} 
			&                                      & \textbf{Top-1} & \multicolumn{1}{c|}{\textbf{Top-5}} & \textbf{Top-1} & \multicolumn{1}{c|}{\textbf{Top-5}} & \textbf{Top-1} & \multicolumn{1}{c|}{\textbf{Top-5}} & \textbf{Top-1} & \multicolumn{1}{c|}{\textbf{Top-5}} & \textbf{Top-1} & \multicolumn{1}{c|}{\textbf{Top-5}} & \textbf{Top-1}         & \textbf{Top-5}        & \textbf{Top-1}            & \textbf{Top-5}            \\ \midrule \midrule
			\multirow{10}{*}{ArcFace}                                                           & GAP~\cite{poursaeed2018generative}                                  & 35.0           & \multicolumn{1}{c|}{25.4}           & 21.3           & \multicolumn{1}{c|}{14.2}           & 25.0           & \multicolumn{1}{c|}{16.5}           & 35.7           & \multicolumn{1}{c|}{22.9}           & 32.5           & \multicolumn{1}{c|}{21.1}           & 15.8                   & 10.1                  & 27.6                      & 18.4                      \\
			& AdvFaces+                            & 75.1           & \multicolumn{1}{c|}{65.7}           & 69.3           & \multicolumn{1}{c|}{60.7}           & 73.0           & \multicolumn{1}{c|}{64.5}           & 77.0           & \multicolumn{1}{c|}{65.9}           & 74.5           & \multicolumn{1}{c|}{63.4}           & 71.2                   & 60.8                  & 73.3                      & 63.5                      \\ \cline{2-16} 
			& FI-UAP~\cite{zhong2022opom}                               & 78.3           & \multicolumn{1}{c|}{70.0}           & 66.1           & \multicolumn{1}{c|}{57.0}           & 74.3           & \multicolumn{1}{c|}{63.9}           & 57.7           & \multicolumn{1}{c|}{43.7}           & 68.6           & \multicolumn{1}{c|}{56.7}           & 50.6                   & 38.3                  & 65.9                      & 54.9                      \\
			& \textbf{AdvCloak}             & \textbf{80.1}  & \multicolumn{1}{c|}{\textbf{71.0}}  & \textbf{74.7}  & \multicolumn{1}{c|}{\textbf{65.7}}  & \textbf{78.0}  & \multicolumn{1}{c|}{\textbf{69.7}}  & \textbf{82.0}  & \multicolumn{1}{c|}{\textbf{71.9}}  & \textbf{78.1}  & \multicolumn{1}{c|}{\textbf{67.4}}  & \textbf{77.4}          & \textbf{67.5}         & \textbf{78.4}             & \textbf{68.9}             \\ \cline{2-16} 
			& OPOM-AffineHull~\cite{zhong2022opom}                      & 79.0           & \multicolumn{1}{c|}{70.4}           & 66.8           & \multicolumn{1}{c|}{57.6}           & 73.0           & \multicolumn{1}{c|}{64.7}           & 59.7           & \multicolumn{1}{c|}{45.1}           & 69.4           & \multicolumn{1}{c|}{58.0}           & 51.7                   & 39.3                  & 66.6                      & 55.8                      \\
			& \textbf{AdvCloak-AffineHull}  & \textbf{79.5}  & \multicolumn{1}{c|}{\textbf{70.6}}  & \textbf{74.1}  & \multicolumn{1}{c|}{\textbf{65.1}}  & \textbf{77.5}  & \multicolumn{1}{c|}{\textbf{69.2}}  & \textbf{81.9}  & \multicolumn{1}{c|}{\textbf{71.7}}  & \textbf{77.9}  & \multicolumn{1}{c|}{\textbf{67.4}}  & \textbf{77.2}          & \textbf{67.2}         & \textbf{78.0}             & \textbf{68.5}             \\ \cline{2-16} 
			& OPOM-ClassCenter~\cite{zhong2022opom}                      & \textbf{82.2}  & \multicolumn{1}{c|}{\textbf{75.1}}  & 70.6           & \multicolumn{1}{c|}{62.1}           & 76.6           & \multicolumn{1}{c|}{69.6}           & 63.3           & \multicolumn{1}{c|}{49.4}           & 73.1           & \multicolumn{1}{c|}{62.8}           & 55.4                   & 43.3                  & 70.2                      & 60.4                      \\
			& \textbf{AdvCloak-ClassCenter} & 81.1           & \multicolumn{1}{c|}{72.3}           & \textbf{75.8}  & \multicolumn{1}{c|}{\textbf{67.5}}  & \textbf{79.0}  & \multicolumn{1}{c|}{\textbf{71.1}}  & \textbf{83.8}  & \multicolumn{1}{c|}{\textbf{74.7}}  & \textbf{80.2}  & \multicolumn{1}{c|}{\textbf{70.9}}  & \textbf{79.0}          & \textbf{69.4}         & \textbf{79.8}             & \textbf{71.0}             \\ \cline{2-16} 
			& OPOM-ConvexHull~\cite{zhong2022opom}                      & \textbf{83.0}  & \multicolumn{1}{c|}{\textbf{75.7}}  & 71.3           & \multicolumn{1}{c|}{63.3}           & 77.9           & \multicolumn{1}{c|}{70.5}           & 64.4           & \multicolumn{1}{c|}{50.0}           & 74.3           & \multicolumn{1}{c|}{64.0}           & 56.8                   & 44.5                  & 71.3                      & 61.3                      \\
			& \textbf{AdvCloak-ConvexHull}  & 81.1           & \multicolumn{1}{c|}{72.7}           & \textbf{75.9}  & \multicolumn{1}{c|}{\textbf{67.5}}  & \textbf{79.1}  & \multicolumn{1}{c|}{\textbf{71.3}}  & \textbf{83.7}  & \multicolumn{1}{c|}{\textbf{74.3}}  & \textbf{79.8}  & \multicolumn{1}{c|}{\textbf{70.2}}  & \textbf{79.0}          & \textbf{69.4}         & \textbf{79.8}             & \textbf{70.9}             \\ \midrule \midrule
			\multirow{10}{*}{CosFace}                                                           & GAP~\cite{poursaeed2018generative}                                  & 42.3           & \multicolumn{1}{c|}{30.6}           & 25.4           & \multicolumn{1}{c|}{16.5}           & 23.7           & \multicolumn{1}{c|}{16.3}           & 29.6           & \multicolumn{1}{c|}{18.0}           & 31.7           & \multicolumn{1}{c|}{21.1}           & 16.8                   & 10.1                  & 28.2                      & 18.8                      \\
			& AdvFaces+                            & 73.3           & \multicolumn{1}{c|}{63.1}           & 70.0           & \multicolumn{1}{c|}{61.4}           & 72.1           & \multicolumn{1}{c|}{63.3}           & 72.0           & \multicolumn{1}{c|}{59.3}           & 70.2           & \multicolumn{1}{c|}{58.0}           & 67.0                   & 56.6                  & 70.8                      & 60.3                      \\ \cline{2-16} 
			& FI-UAP~\cite{zhong2022opom}                               & 75.1           & \multicolumn{1}{c|}{65.8}           & 67.0           & \multicolumn{1}{c|}{57.9}           & 70.7           & \multicolumn{1}{c|}{62.0}           & 48.6           & \multicolumn{1}{c|}{33.4}           & 60.0           & \multicolumn{1}{c|}{46.4}           & 41.5                   & 29.6                  & 60.5                      & 49.2                      \\
			& \textbf{AdvCloak}             & \textbf{78.4}  & \multicolumn{1}{c|}{\textbf{68.5}}  & \textbf{74.3}  & \multicolumn{1}{c|}{\textbf{65.4}}  & \textbf{76.5}  & \multicolumn{1}{c|}{\textbf{68.1}}  & \textbf{76.6}  & \multicolumn{1}{c|}{\textbf{64.9}}  & \textbf{73.6}  & \multicolumn{1}{c|}{\textbf{61.4}}  & \textbf{72.1}          & \textbf{61.2}         & \textbf{75.3}             & \textbf{64.9}             \\ \cline{2-16} 
			& OPOM-AffineHull~\cite{zhong2022opom}                      & 76.1           & \multicolumn{1}{c|}{67.4}           & 68.8           & \multicolumn{1}{c|}{60.5}           & 72.5           & \multicolumn{1}{c|}{63.9}           & 50.2           & \multicolumn{1}{c|}{35.2}           & 61.4           & \multicolumn{1}{c|}{49.3}           & 43.7                   & 31.8                  & 62.1                      & 51.3                      \\
			& \textbf{AdvCloak-AffineHull}  & \textbf{77.7}  & \multicolumn{1}{c|}{\textbf{68.1}}  & \textbf{74.0}  & \multicolumn{1}{c|}{\textbf{65.3}}  & \textbf{76.3}  & \multicolumn{1}{c|}{\textbf{67.7}}  & \textbf{76.7}  & \multicolumn{1}{c|}{\textbf{64.7}}  & \textbf{73.5}  & \multicolumn{1}{c|}{\textbf{61.2}}  & \textbf{72.1}          & \textbf{61.4}         & \textbf{75.0}             & \textbf{64.7}             \\ \cline{2-16} 
			& OPOM-ClassCenter~\cite{zhong2022opom}                      & \textbf{80.1}  & \multicolumn{1}{c|}{\textbf{71.8}}  & 72.6           & \multicolumn{1}{c|}{64.9}           & 76.5           & \multicolumn{1}{c|}{69.1}           & 53.6           & \multicolumn{1}{c|}{38.9}           & 66.2           & \multicolumn{1}{c|}{53.5}           & 47.2                   & 34.9                  & 66.0                      & 55.5                      \\
			& \textbf{AdvCloak-ClassCenter} & 78.7           & \multicolumn{1}{c|}{69.5}           & \textbf{75.4}  & \multicolumn{1}{c|}{\textbf{67.4}}  & \textbf{77.5}  & \multicolumn{1}{c|}{\textbf{69.6}}  & \textbf{78.6}  & \multicolumn{1}{c|}{\textbf{68.0}}  & \textbf{76.0}  & \multicolumn{1}{c|}{\textbf{64.8}}  & \textbf{74.4}          & \textbf{64.3}         & \textbf{76.8}             & \textbf{67.3}             \\ \cline{2-16} 
			& OPOM-ConvexHull~\cite{zhong2022opom}                      & \textbf{81.4}  & \multicolumn{1}{c|}{\textbf{73.2}}  & 73.9           & \multicolumn{1}{c|}{66.0}           & \textbf{77.6}  & \multicolumn{1}{c|}{\textbf{70.7}}  & 54.8           & \multicolumn{1}{c|}{39.6}           & 67.3           & \multicolumn{1}{c|}{54.7}           & 48.4                   & 36.3                  & 67.2                      & 56.7                      \\
			& \textbf{AdvCloak-ConvexHull}  & 79.1           & \multicolumn{1}{c|}{69.7}           & \textbf{75.7}  & \multicolumn{1}{c|}{\textbf{67.0}}  & 77.4           & \multicolumn{1}{c|}{69.6}           & \textbf{78.1}  & \multicolumn{1}{c|}{\textbf{66.8}}  & \textbf{75.2}  & \multicolumn{1}{c|}{\textbf{63.6}}  & \textbf{73.6}          & \textbf{63.0}         & \textbf{76.5}             & \textbf{66.6}             \\ \midrule \midrule
			\multirow{10}{*}{Softmax}                                                           & GAP~\cite{poursaeed2018generative}                                  & 17.8           & \multicolumn{1}{c|}{11.4}           & 11.1           & \multicolumn{1}{c|}{7.6}            & 12.5           & \multicolumn{1}{c|}{7.6}            & 21.7           & \multicolumn{1}{c|}{12.0}           & 18.8           & \multicolumn{1}{c|}{11.1}           & 7.0                    & 3.8                   & 14.8                      & 8.9                       \\
			& AdvFaces+                            & 75.3           & \multicolumn{1}{c|}{65.5}           & 70.0           & \multicolumn{1}{c|}{61.5}           & 73.4           & \multicolumn{1}{c|}{65.0}           & 80.7           & \multicolumn{1}{c|}{71.3}           & 77.0           & \multicolumn{1}{c|}{67.2}           & 72.1                   & 61.8                  & 74.7                      & 65.4                      \\ \cline{2-16} 
			& FI-UAP~\cite{zhong2022opom}                               & 65.2           & \multicolumn{1}{c|}{53.6}           & 55.7           & \multicolumn{1}{c|}{45.4}           & 63.1           & \multicolumn{1}{c|}{53.3}           & 66.7           & \multicolumn{1}{c|}{53.2}           & 70.5           & \multicolumn{1}{c|}{59.6}           & 44.3                   & 33.0                  & 60.9                      & 49.7                      \\
			& \textbf{AdvCloak (Ours)}             & \textbf{79.0}  & \multicolumn{1}{c|}{\textbf{69.3}}  & \textbf{73.8}  & \multicolumn{1}{c|}{\textbf{64.6}}  & \textbf{77.9}  & \multicolumn{1}{c|}{\textbf{69.6}}  & \textbf{84.0}  & \multicolumn{1}{c|}{\textbf{75.0}}  & \textbf{79.7}  & \multicolumn{1}{c|}{\textbf{70.5}}  & \textbf{77.2}          & \textbf{66.3}         & \textbf{78.6}             & \textbf{69.2}             \\ \cline{2-16} 
			& OPOM-AffineHull~\cite{zhong2022opom}                      & 65.9           & \multicolumn{1}{c|}{54.6}           & 56.1           & \multicolumn{1}{c|}{46.0}           & 63.8           & \multicolumn{1}{c|}{53.4}           & 67.8           & \multicolumn{1}{c|}{54.5}           & 71.7           & \multicolumn{1}{c|}{60.3}           & 44.0                   & 32.8                  & 61.5                      & 50.3                      \\
			& \textbf{AdvCloak-AffineHull}  & \textbf{79.2}  & \multicolumn{1}{c|}{\textbf{69.8}}  & \textbf{74.1}  & \multicolumn{1}{c|}{\textbf{64.8}}  & \textbf{78.0}  & \multicolumn{1}{c|}{\textbf{69.5}}  & \textbf{83.7}  & \multicolumn{1}{c|}{\textbf{75.1}}  & \textbf{79.9}  & \multicolumn{1}{c|}{\textbf{70.7}}  & \textbf{76.8}          & \textbf{66.3}         & \textbf{78.6}             & \textbf{69.4}             \\ \cline{2-16} 
			& OPOM-ClassCenter~\cite{zhong2022opom}                      & 69.0           & \multicolumn{1}{c|}{58.7}           & 60.7           & \multicolumn{1}{c|}{51.0}           & 68.0           & \multicolumn{1}{c|}{59.0}           & 71.7           & \multicolumn{1}{c|}{59.5}           & 75.5           & \multicolumn{1}{c|}{65.3}           & 48.8                   & 37.3                  & 65.6                      & 55.2                      \\
			& \textbf{AdvCloak-ClassCenter} & \textbf{81.6}  & \multicolumn{1}{c|}{\textbf{73.0}}  & \textbf{76.9}  & \multicolumn{1}{c|}{\textbf{68.6}}  & \textbf{80.5}  & \multicolumn{1}{c|}{\textbf{72.3}}  & \textbf{87.3}  & \multicolumn{1}{c|}{\textbf{79.3}}  & \textbf{83.2}  & \multicolumn{1}{c|}{\textbf{75.6}}  & \textbf{80.4}          & \textbf{70.9}         & \textbf{81.7}             & \textbf{73.3}             \\ \cline{2-16} 
			& OPOM-ConvexHull~\cite{zhong2022opom}                      & 70.7           & \multicolumn{1}{c|}{60.2}           & 61.8           & \multicolumn{1}{c|}{52.1}           & 69.4           & \multicolumn{1}{c|}{60.5}           & 72.6           & \multicolumn{1}{c|}{60.7}           & 76.2           & \multicolumn{1}{c|}{66.4}           & 50.2                   & 38.7                  & 66.8                      & 56.4                      \\
			& \textbf{AdvCloak-ConvexHull}  & \textbf{80.7}  & \multicolumn{1}{c|}{\textbf{71.7}}  & \textbf{75.3}  & \multicolumn{1}{c|}{\textbf{66.8}}  & \textbf{79.6}  & \multicolumn{1}{c|}{\textbf{71.2}}  & \textbf{86.1}  & \multicolumn{1}{c|}{\textbf{78.2}}  & \textbf{82.7}  & \multicolumn{1}{c|}{\textbf{74.1}}  & \textbf{79.1}          & \textbf{69.3}         & \textbf{80.6}             & \textbf{71.9}             \\ \bottomrule
		\end{tabular}
	}
\end{table*}

\begin{table*}[]
\setlength\tabcolsep{2.5pt}
	\renewcommand{\arraystretch}{1.4}
	\caption{\centering Top-1 and Top-5 protection success rates (\%) are reported against black-box models under the 1:N identification setting on the Privacy-Celebrities dataset. The AdvCloak-based methods are our proposed methods.}
	\centering
	\label{ten_ms}
	\vspace{-2mm}
	\resizebox{\linewidth}{!}{
		\begin{tabular}{c|c|cccccccccccc|cc}
			\toprule
			\multirow{3}{*}{\textbf{\begin{tabular}[c]{@{}c@{}}Source\\ Model\end{tabular}}} & \multirow{3}{*}{\textbf{Method}}     & \multicolumn{12}{c|}{\textbf{Target Model}}                                                                                                                                                                                                                                                                                       & \multicolumn{2}{c}{\multirow{2}{*}{\textbf{Average}}} \\ \cline{3-14}
			&                                      & \multicolumn{2}{c|}{\textbf{ArcFace}}                & \multicolumn{2}{c|}{\textbf{CosFace}}                & \multicolumn{2}{c|}{\textbf{SFace}}                  & \multicolumn{2}{c|}{\textbf{MobileNet}}              & \multicolumn{2}{c|}{\textbf{SENet}}                  & \multicolumn{2}{c|}{\textbf{IncRes}} & \multicolumn{2}{c}{}                                  \\ \cline{15-16} 
			&                                      & \textbf{Top-1} & \multicolumn{1}{c|}{\textbf{Top-5}} & \textbf{Top-1} & \multicolumn{1}{c|}{\textbf{Top-5}} & \textbf{Top-1} & \multicolumn{1}{c|}{\textbf{Top-5}} & \textbf{Top-1} & \multicolumn{1}{c|}{\textbf{Top-5}} & \textbf{Top-1} & \multicolumn{1}{c|}{\textbf{Top-5}} & \textbf{Top-1}         & \textbf{Top-5}        & \textbf{Top-1}            & \textbf{Top-5}            \\ \midrule \midrule
			\multirow{10}{*}{ArcFace}                                                           & GAP~\cite{poursaeed2018generative}                                  & 40.8           & \multicolumn{1}{c|}{30.9}           & 29.4           & \multicolumn{1}{c|}{21.5}           & 34.8           & \multicolumn{1}{c|}{26.3}           & 45.6           & \multicolumn{1}{c|}{33.6}           & 39.0           & \multicolumn{1}{c|}{28.3}           & 26.0                   & 17.0                  & 35.9                      & 26.3                      \\
			& AdvFaces+                            & 58.4           & \multicolumn{1}{c|}{46.6}           & 51.8           & \multicolumn{1}{c|}{40.6}           & 56.2           & \multicolumn{1}{c|}{45.3}           & 62.3           & \multicolumn{1}{c|}{47.6}           & 58.8           & \multicolumn{1}{c|}{45.0}           & 52.1                   & 38.0                  & 56.6                      & 43.9                      \\ \cline{2-16} 
			& FI-UAP~\cite{zhong2022opom}                               & 61.9           & \multicolumn{1}{c|}{51.2}           & 47.7           & \multicolumn{1}{c|}{37.4}           & 55.3           & \multicolumn{1}{c|}{45.5}           & 50.9           & \multicolumn{1}{c|}{36.2}           & 54.3           & \multicolumn{1}{c|}{41.2}           & 36.2                   & 25.4                  & 51.0                      & 39.5                      \\
			& \textbf{AdvCloak}             & \textbf{67.7}  & \multicolumn{1}{c|}{\textbf{56.9}}  & \textbf{60.4}  & \multicolumn{1}{c|}{\textbf{50.4}}  & \textbf{64.6}  & \multicolumn{1}{c|}{\textbf{55.1}}  & \textbf{71.8}  & \multicolumn{1}{c|}{\textbf{60.1}}  & \textbf{67.2}  & \multicolumn{1}{c|}{\textbf{55.0}}  & \textbf{62.4}          & \textbf{50.3}         & \textbf{65.7}             & \textbf{54.6}             \\ \cline{2-16} 
			& OPOM-AffineHull~\cite{zhong2022opom}                      & 64.9           & \multicolumn{1}{c|}{54.6}           & 50.6           & \multicolumn{1}{c|}{40.8}           & 58.1           & \multicolumn{1}{c|}{48.9}           & 53.3           & \multicolumn{1}{c|}{38.5}           & 57.2           & \multicolumn{1}{c|}{44.4}           & 38.6                   & 27.3                  & 53.8                      & 42.4                      \\
			& \textbf{AdvCloak-AffineHull}  & \textbf{66.8}  & \multicolumn{1}{c|}{\textbf{55.9}}  & \textbf{59.8}  & \multicolumn{1}{c|}{\textbf{49.2}}  & \textbf{63.6}  & \multicolumn{1}{c|}{\textbf{53.9}}  & \textbf{70.3}  & \multicolumn{1}{c|}{\textbf{57.8}}  & \textbf{65.8}  & \multicolumn{1}{c|}{\textbf{53.2}}  & \textbf{61.1}          & \textbf{48.6}         & \textbf{64.6}             & \textbf{53.1}             \\ \cline{2-16} 
			& OPOM-ClassCenter~\cite{zhong2022opom}                      & 66.9           & \multicolumn{1}{c|}{\textbf{57.5}}  & 53.8           & \multicolumn{1}{c|}{44.5}           & 61.7           & \multicolumn{1}{c|}{53.0}           & 55.1           & \multicolumn{1}{c|}{41.1}           & 60.4           & \multicolumn{1}{c|}{48.1}           & 42.0                   & 30.5                  & 56.7                      & 45.8                      \\
			& \textbf{AdvCloak-ClassCenter} & \textbf{67.6}  & \multicolumn{1}{c|}{56.9}           & \textbf{61.0}  & \multicolumn{1}{c|}{\textbf{50.9}}  & \textbf{64.9}  & \multicolumn{1}{c|}{\textbf{55.3}}  & \textbf{72.3}  & \multicolumn{1}{c|}{\textbf{60.0}}  & \textbf{67.9}  & \multicolumn{1}{c|}{\textbf{55.8}}  & \textbf{62.5}          & \textbf{50.4}         & \textbf{66.0}             & \textbf{54.9}             \\ \cline{2-16} 
			& OPOM-ConvexHull~\cite{zhong2022opom}                      & \textbf{68.5}  & \multicolumn{1}{c|}{\textbf{59.1}}  & 55.0           & \multicolumn{1}{c|}{45.9}           & 63.0           & \multicolumn{1}{c|}{54.0}           & 56.4           & \multicolumn{1}{c|}{41.9}           & 61.7           & \multicolumn{1}{c|}{49.4}           & 43.3                   & 31.7                  & 58.0                      & 47.0                      \\
			& \textbf{AdvCloak-ConvexHull}  & \textbf{68.5}  & \multicolumn{1}{c|}{57.6}           & \textbf{61.4}  & \multicolumn{1}{c|}{\textbf{51.3}}  & \textbf{65.3}  & \multicolumn{1}{c|}{\textbf{55.8}}  & \textbf{72.3}  & \multicolumn{1}{c|}{\textbf{60.3}}  & \textbf{67.8}  & \multicolumn{1}{c|}{\textbf{55.8}}  & \textbf{62.9}          & \textbf{50.7}         & \textbf{66.4}             & \textbf{55.2}             \\ \midrule \midrule
			\multirow{10}{*}{CosFace}                                                           & GAP~\cite{poursaeed2018generative}                                  & 42.5           & \multicolumn{1}{c|}{32.6}           & 30.4           & \multicolumn{1}{c|}{21.7}           & 32.1           & \multicolumn{1}{c|}{23.1}           & 39.6           & \multicolumn{1}{c|}{26.6}           & 37.4           & \multicolumn{1}{c|}{26.9}           & 17.2                   & 9.5                   & 33.2                      & 23.4                      \\
			& AdvFaces+                            & 57.9           & \multicolumn{1}{c|}{44.7}           & 52.1           & \multicolumn{1}{c|}{40.9}           & 55.6           & \multicolumn{1}{c|}{44.1}           & 58.1           & \multicolumn{1}{c|}{43.1}           & 54.8           & \multicolumn{1}{c|}{40.4}           & 49.8                   & 35.3                  & 54.7                      & 41.4                      \\ \cline{2-16} 
			& FI-UAP~\cite{zhong2022opom}                               & 62.1           & \multicolumn{1}{c|}{50.8}           & 50.4           & \multicolumn{1}{c|}{40.3}           & 55.9           & \multicolumn{1}{c|}{46.1}           & 46.1           & \multicolumn{1}{c|}{31.4}           & 48.8           & \multicolumn{1}{c|}{36.0}           & 32.5                   & 21.2                  & 49.3                      & 37.6                      \\
			& \textbf{AdvCloak}             & \textbf{67.3}  & \multicolumn{1}{c|}{\textbf{56.5}}  & \textbf{60.8}  & \multicolumn{1}{c|}{\textbf{50.2}}  & \textbf{64.4}  & \multicolumn{1}{c|}{\textbf{54.1}}  & \textbf{69.2}  & \multicolumn{1}{c|}{\textbf{56.5}}  & \textbf{64.1}  & \multicolumn{1}{c|}{\textbf{51.2}}  & \textbf{58.5}          & \textbf{45.2}         & \textbf{64.0}             & \textbf{52.3}             \\ \cline{2-16} 
			& OPOM-AffineHull~\cite{zhong2022opom}                      & 65.0           & \multicolumn{1}{c|}{54.3}           & 54.2           & \multicolumn{1}{c|}{43.9}           & 59.7           & \multicolumn{1}{c|}{49.9}           & 49.1           & \multicolumn{1}{c|}{34.2}           & 52.6           & \multicolumn{1}{c|}{39.5}           & 34.8                   & 23.4                  & 52.6                      & 40.9                      \\
			& \textbf{AdvCloak-AffineHull}  & \textbf{66.7}  & \multicolumn{1}{c|}{\textbf{55.4}}  & \textbf{60.5}  & \multicolumn{1}{c|}{\textbf{49.9}}  & \textbf{64.2}  & \multicolumn{1}{c|}{\textbf{53.4}}  & \textbf{68.3}  & \multicolumn{1}{c|}{\textbf{55.2}}  & \textbf{63.4}  & \multicolumn{1}{c|}{\textbf{50.4}}  & \textbf{57.9}          & \textbf{44.4}         & \textbf{63.5}             & \textbf{51.4}             \\ \cline{2-16} 
			& OPOM-ClassCenter~\cite{zhong2022opom}                      & 66.0           & \multicolumn{1}{c|}{\textbf{56.2}}  & 56.6           & \multicolumn{1}{c|}{46.5}           & 61.7           & \multicolumn{1}{c|}{52.5}           & 49.5           & \multicolumn{1}{c|}{35.0}           & 54.3           & \multicolumn{1}{c|}{41.4}           & 36.4                   & 25.1                  & 54.1                      & 42.8                      \\
			& \textbf{AdvCloak-ClassCenter} & \textbf{66.5}  & \multicolumn{1}{c|}{54.9}           & \textbf{61.1}  & \multicolumn{1}{c|}{\textbf{50.3}}  & \textbf{64.3}  & \multicolumn{1}{c|}{\textbf{54.1}}  & \textbf{68.2}  & \multicolumn{1}{c|}{\textbf{55.1}}  & \textbf{64.7}  & \multicolumn{1}{c|}{\textbf{51.7}}  & \textbf{59.1}          & \textbf{45.4}         & \textbf{64.0}             & \textbf{51.9}             \\ \cline{2-16} 
			& OPOM-ConvexHull~\cite{zhong2022opom}                      & \textbf{67.4}  & \multicolumn{1}{c|}{\textbf{58.0}}  & 57.9           & \multicolumn{1}{c|}{47.9}           & 63.2           & \multicolumn{1}{c|}{54.1}           & 50.8           & \multicolumn{1}{c|}{35.9}           & 56.1           & \multicolumn{1}{c|}{43.2}           & 37.8                   & 26.3                  & 55.5                      & 44.2                      \\
			& \textbf{AdvCloak-ConvexHull}  & \textbf{67.4}  & \multicolumn{1}{c|}{56.5}           & \textbf{61.6}  & \multicolumn{1}{c|}{\textbf{51.0}}  & \textbf{64.6}  & \multicolumn{1}{c|}{\textbf{54.6}}  & \textbf{69.2}  & \multicolumn{1}{c|}{\textbf{55.9}}  & \textbf{65.0}  & \multicolumn{1}{c|}{\textbf{51.9}}  & \textbf{59.1}          & \textbf{45.8}         & \textbf{64.5}             & \textbf{52.6}             \\ \midrule \midrule
			\multirow{10}{*}{Softmax}                                                           & GAP~\cite{poursaeed2018generative}                                  & 30.3           & \multicolumn{1}{c|}{21.0}           & 21.0           & \multicolumn{1}{c|}{13.3}           & 24.1           & \multicolumn{1}{c|}{16.3}           & 43.4           & \multicolumn{1}{c|}{31.2}           & 33.2           & \multicolumn{1}{c|}{23.0}           & 12.8                   & 6.6                   & 27.5                      & 18.5                      \\
			& AdvFaces+                            & 60.1           & \multicolumn{1}{c|}{48.2}           & 53.5           & \multicolumn{1}{c|}{42.4}           & 58.1           & \multicolumn{1}{c|}{47.2}           & 68.5           & \multicolumn{1}{c|}{55.7}           & 63.4           & \multicolumn{1}{c|}{50.4}           & 53.8                   & 41.0                  & 59.6                      & 47.5                      \\ \cline{2-16} 
			& FI-UAP~\cite{zhong2022opom}                               & 51.6           & \multicolumn{1}{c|}{40.6}           & 40.4           & \multicolumn{1}{c|}{30.6}           & 47.9           & \multicolumn{1}{c|}{38.1}           & 55.6           & \multicolumn{1}{c|}{42.2}           & 56.1           & \multicolumn{1}{c|}{44.4}           & 33.3                   & 22.8                  & 47.5                      & 36.4                      \\
			& \textbf{AdvCloak}             & \textbf{66.1}  & \multicolumn{1}{c|}{\textbf{54.4}}  & \textbf{59.7}  & \multicolumn{1}{c|}{\textbf{49.2}}  & \textbf{63.7}  & \multicolumn{1}{c|}{\textbf{53.5}}  & \textbf{72.3}  & \multicolumn{1}{c|}{\textbf{60.0}}  & \textbf{67.4}  & \multicolumn{1}{c|}{\textbf{55.3}}  & \textbf{60.0}          & \textbf{47.2}         & \textbf{64.9}             & \textbf{53.3}             \\ \cline{2-16} 
			& OPOM-AffineHull~\cite{zhong2022opom}                      & 53.3           & \multicolumn{1}{c|}{42.7}           & 42.6           & \multicolumn{1}{c|}{32.6}           & 49.7           & \multicolumn{1}{c|}{40.1}           & 58.0           & \multicolumn{1}{c|}{44.5}           & 58.5           & \multicolumn{1}{c|}{46.4}           & 35.1                   & 23.8                  & 49.5                      & 38.3                      \\
			& \textbf{AdvCloak-AffineHull}  & \textbf{65.4}  & \multicolumn{1}{c|}{\textbf{54.0}}  & \textbf{59.4}  & \multicolumn{1}{c|}{\textbf{48.8}}  & \textbf{63.0}  & \multicolumn{1}{c|}{\textbf{53.0}}  & \textbf{71.7}  & \multicolumn{1}{c|}{\textbf{59.3}}  & \textbf{66.6}  & \multicolumn{1}{c|}{\textbf{53.8}}  & \textbf{59.2}          & \textbf{45.9}         & \textbf{64.2}             & \textbf{52.5}             \\ \cline{2-16} 
			& OPOM-ClassCenter~\cite{zhong2022opom}                      & 56.3           & \multicolumn{1}{c|}{45.7}           & 46.3           & \multicolumn{1}{c|}{36.7}           & 53.9           & \multicolumn{1}{c|}{43.7}           & 61.9           & \multicolumn{1}{c|}{49.2}           & 62.1           & \multicolumn{1}{c|}{51.1}           & 37.4                   & 26.1                  & 53.0                      & 42.1                      \\
			& \textbf{AdvCloak-ClassCenter} & \textbf{68.3}  & \multicolumn{1}{c|}{\textbf{57.6}}  & \textbf{62.1}  & \multicolumn{1}{c|}{\textbf{51.8}}  & \textbf{66.1}  & \multicolumn{1}{c|}{\textbf{56.4}}  & \textbf{75.9}  & \multicolumn{1}{c|}{\textbf{65.0}}  & \textbf{71.5}  & \multicolumn{1}{c|}{\textbf{60.0}}  & \textbf{63.0}          & \textbf{50.5}         & \textbf{67.8}             & \textbf{56.9}             \\ \cline{2-16} 
			& OPOM-ConvexHull~\cite{zhong2022opom}                      & 58.3           & \multicolumn{1}{c|}{47.8}           & 48.0           & \multicolumn{1}{c|}{38.4}           & 55.6           & \multicolumn{1}{c|}{45.6}           & 62.8           & \multicolumn{1}{c|}{50.3}           & 63.7           & \multicolumn{1}{c|}{52.5}           & 38.9                   & 27.7                  & 54.6                      & 43.7                      \\
			& \textbf{AdvCloak-ConvexHull}  & \textbf{67.4}  & \multicolumn{1}{c|}{\textbf{56.2}}  & \textbf{61.1}  & \multicolumn{1}{c|}{\textbf{50.2}}  & \textbf{64.7}  & \multicolumn{1}{c|}{\textbf{54.8}}  & \textbf{74.4}  & \multicolumn{1}{c|}{\textbf{62.8}}  & \textbf{69.5}  & \multicolumn{1}{c|}{\textbf{57.3}}  & \textbf{60.9}          & \textbf{48.1}         & \textbf{66.3}             & \textbf{54.9}             \\ \bottomrule
	\end{tabular}}
\end{table*}

\subsection{Effectiveness Of Privacy Protection Performance}
\subsubsection{Attack Performance on Black-Box Models}
We evaluate the proposed AdvCloak-based methods, including AdvCloak, AdvCloak-AffineHull, AdvCloak-ClassCenter, and AdvCloak-ConvexHull, with other comparison methods, \textit{i.e.}, GAP~\cite{poursaeed2018generative}, AdvFaces+, FI-UAP, and OPOM-based methods~\cite{zhong2022opom}, including OPOM-AffineHull, OPOM-ClassCenter, and OPOM-ConvexHull, on the Privacy-Commons dataset and Privacy-Celebrities dataset. Specifically, We first generate person-specific (class-wise) masks against ArcFace, CosFace, and Softmax by the above methods using ten inference images of each person and add each adversarial mask to all the test images of the source identity to form adversarial images (protected images). Then we feed adversarial face images to all the six target black models for testing the protection performance on black-box models.

Top-1 and Top-5 protection success rates (\%) on the Privacy-Commons dataset and Privacy-Celebrities dataset are reported in Table \ref{ten} and Table \ref{ten_ms} respectively. The results show that our proposed AdvCloak outperforms other regular methods, \textit{i.e.}, GAP~\cite{poursaeed2018generative}, and AdvFaces+ in protection performance. In particular, compared with baseline methods including FI-UAP and OPOM~\cite{zhong2022opom}, our method significantly improves the protection performance across all the models on both datasets. For the vanilla privacy masks crafted on the Softmax model, the average protection success rate increases from 60.9\% and 49.3\% to 78.6\% and 64.9\%. It can also be observed that the proposed modeling methods of the feature subspace including ClassCenter and ConvexHull achieve higher protection success rates, which demonstrates that combining with modeling methods can bring better intra-class generalization ability for the adversarial masks.

\begin{figure*}[t]
	\centering
	\subfloat[]{
		\includegraphics[width=0.42\textwidth]{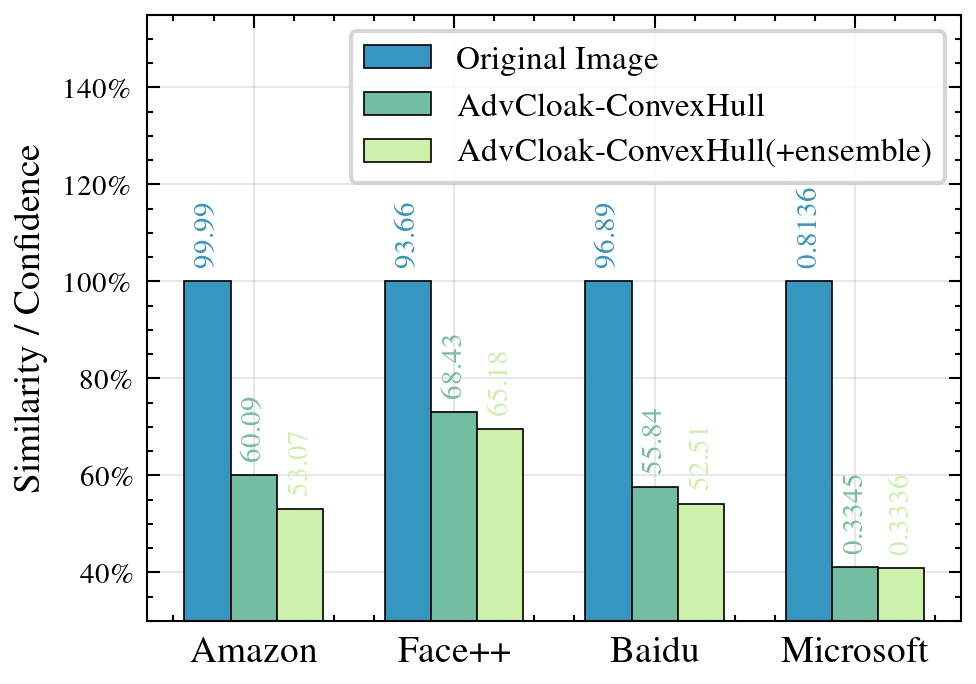}
	}
	\subfloat[]{
		\includegraphics[width=0.43\textwidth]{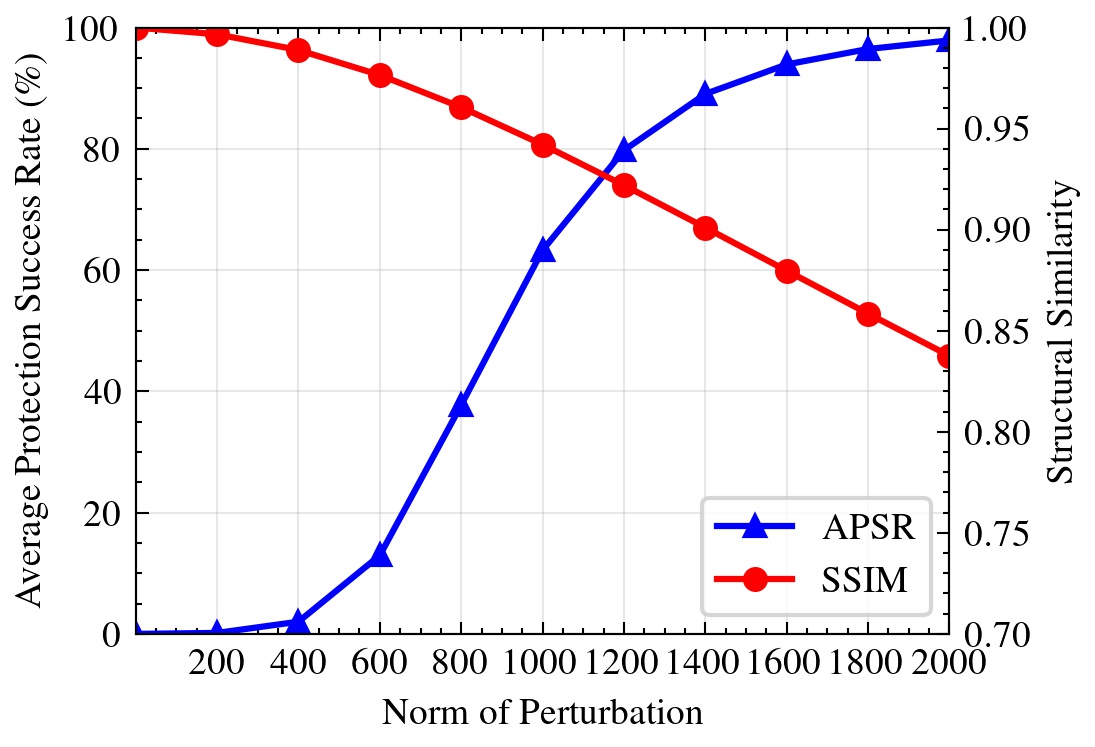}
	}
	
	\subfloat[]{
		\includegraphics[width=0.39\textwidth]{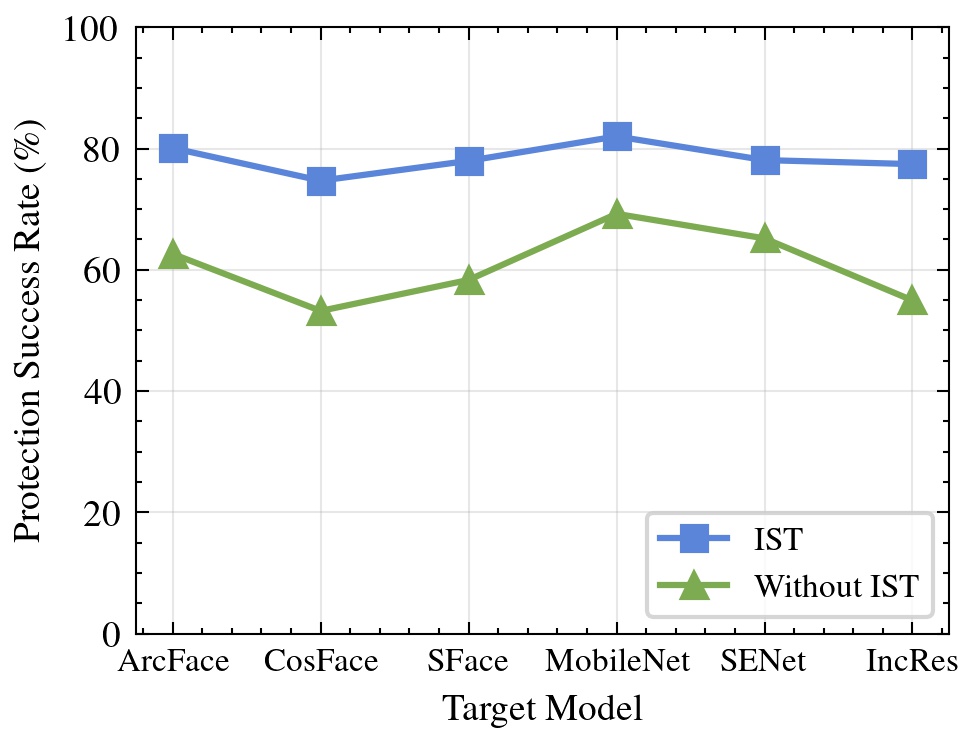}
	}
	\subfloat[]{
		\includegraphics[width=0.51\textwidth]{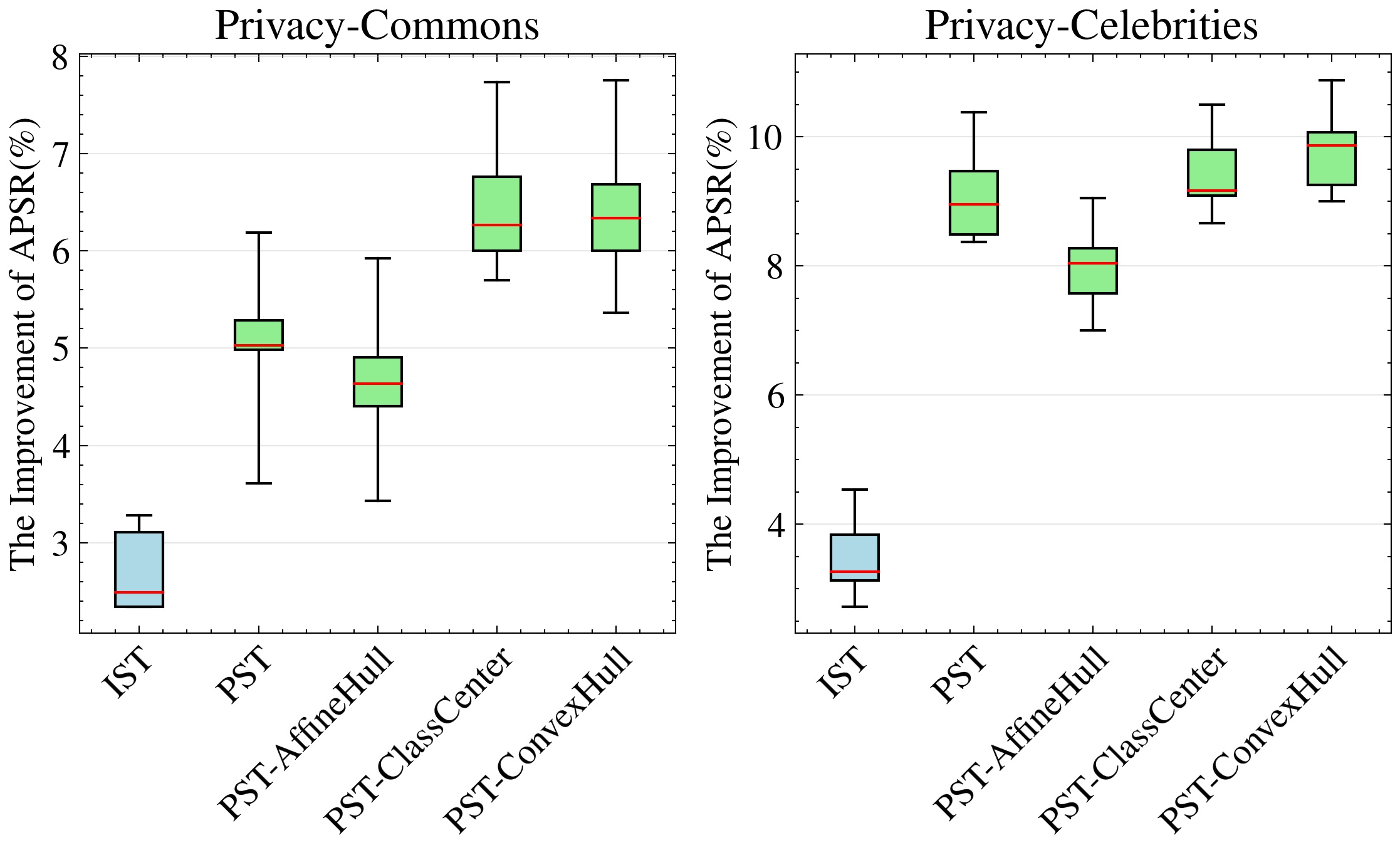}
	}
	
	\caption{(a) The normalized average similarity/confidence scores (lower is better) are reported against Commercial APIs (Amazon~\cite{Amazon}, Face++~\cite{Face++}, Baidu~\cite{baidu}, and Microsoft~\cite{Microsoft}). (b) The trade-off between the average protection success rate and the structural similarity. (c) Ablation study for the image-specific training (IST). (d) Ablation study for the person-specific training. }
	\label{api_aba}
\end{figure*}

\subsubsection{Privacy Protection Performance on APIs}
In Figure \ref{api_aba} (a), we apply our proposed AdvCloak-ConvexHull to test the face privacy protection performance against Commercial APIs (Amazon~\cite{Amazon}, Face++~\cite{Face++}, Baidu~\cite{baidu}, and Microsoft~\cite{Microsoft}). We randomly select 50 people from the Privacy-Commons dataset, each with 5 test images. Considering that the commercial APIs are trained on large datasets, we train the network against two source models trained on CASIA-WebFace and MS-Celeb-1M~\cite{guo2016ms} to increase the transferability to black-box models. It is worth mentioning that the proposed method decreases the similarity score from 0.8136 to 0.3336 when tested on Microsoft APIs, which indicates promising privacy protection in practical applications.

\subsection{Ablation Studies}
\subsubsection{Ablation Study on Perturbation Norm} 
In the above experiments to verify the privacy protection performance of different methods, we normalize each perturbation produced by different methods to the same $l_2$ norm for a fair comparison. In Figure \ref{api_aba} (b), we can observe a trade-off between the protection performance and the structural similarity by projecting the perturbations to different magnitudes. A higher structural similarity value corresponds to superior image quality. Consequently, in our experiments, aiming for a balance between high image quality and robust protection performance, we constrain the $l_2$ norm of all masks generated by different methods to 1,200.

\subsubsection{Effectiveness of Image-specific Training} Here we investigate the effectiveness of the image-specific training (IST) which is used to synthesize adversarial perturbations to adapt to different faces. To facilitate comparison, we train two distinct AdvCloak models against ArcFace, one incorporating IST and the other without.

The experimental results are shown in Figure \ref{api_aba} (c). It is evident that the AdvCloak model trained with image-specific training (IST) surpasses the model without IST across all six black-box models. These results demonstrate the beneficial impact of IST in facilitating the model's adaptation to diverse facial features, thereby ensuring the effectiveness in privacy protection.

\subsubsection{Effectiveness of Person-specific Training} Person-specific training (PST) enhances the optimization of the network through aggregation and generalization. To explore its impact, we conducted an ablation study comparing person-specific training with image-specific training (IST) during the training stage \uppercase\expandafter{\romannumeral2}. We compared the improvement of the average protection success rate brought by the IST, PST, and PST combined with modeling methods of feature subspace, as shown in Figure \ref{api_aba} (d). The results show that PST outperforms IST in enhancing protection performance. This superiority arises from the ability of person-specific training to capture facial salient features and ensure the generated mask's effective generalization to all images of the person, not confined to a specific image. Furthermore, feature subspace modeling for each person contributes to the heightened generalization capability of the adversarial masks.

\begin{figure*}[ht]
	\centering
	\subfloat[]{
		\includegraphics[width=0.45\textwidth]{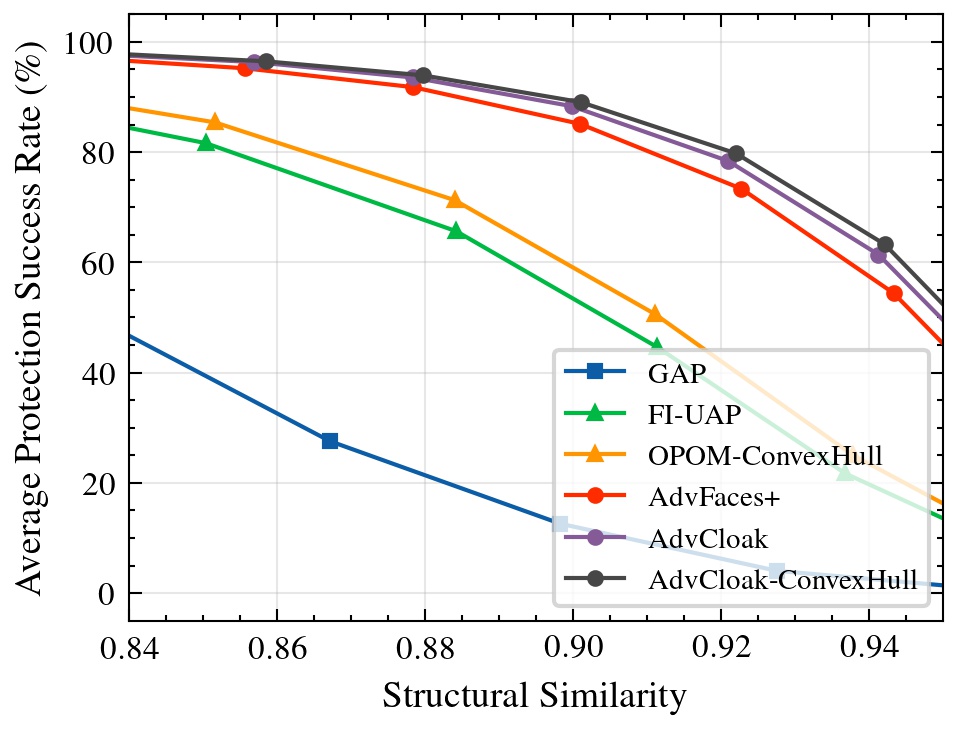}
	}
	\subfloat[]{
		\includegraphics[width=0.45\textwidth]{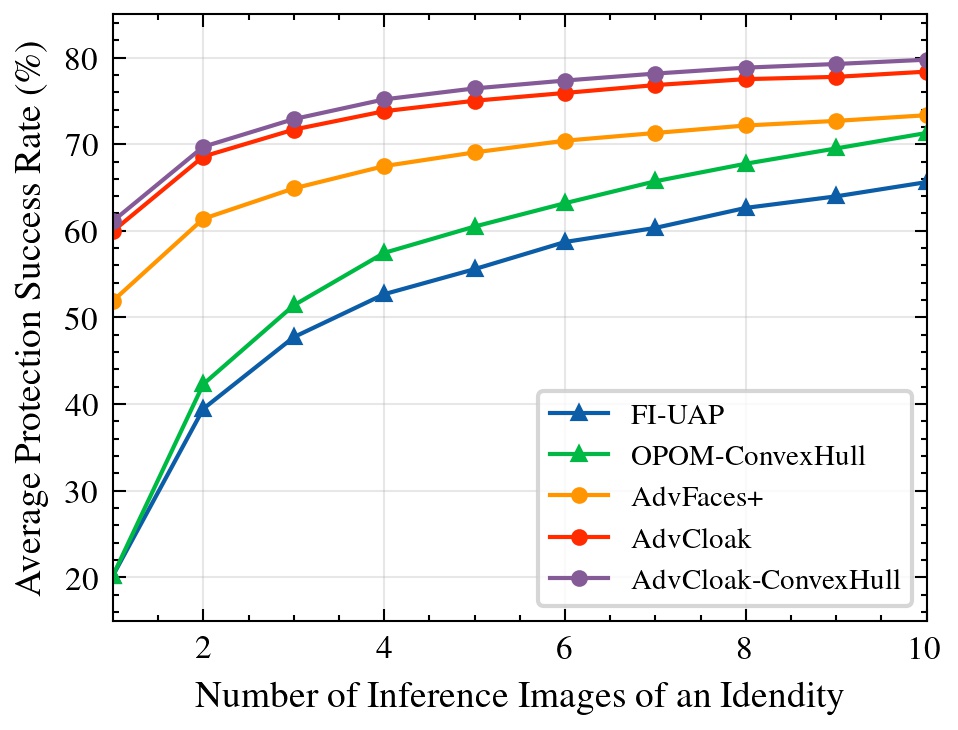}
	}
	\caption{(a) Comparison of the universal privacy masks generated by different methods in terms of the image quality and the average protection success rate. (b) The relationship between the number of inference images and the effectiveness of privacy protection.}
	\label{discuss}
\end{figure*}

\begin{figure*}[!ht]
	\centering
	\includegraphics[width=0.95\linewidth]{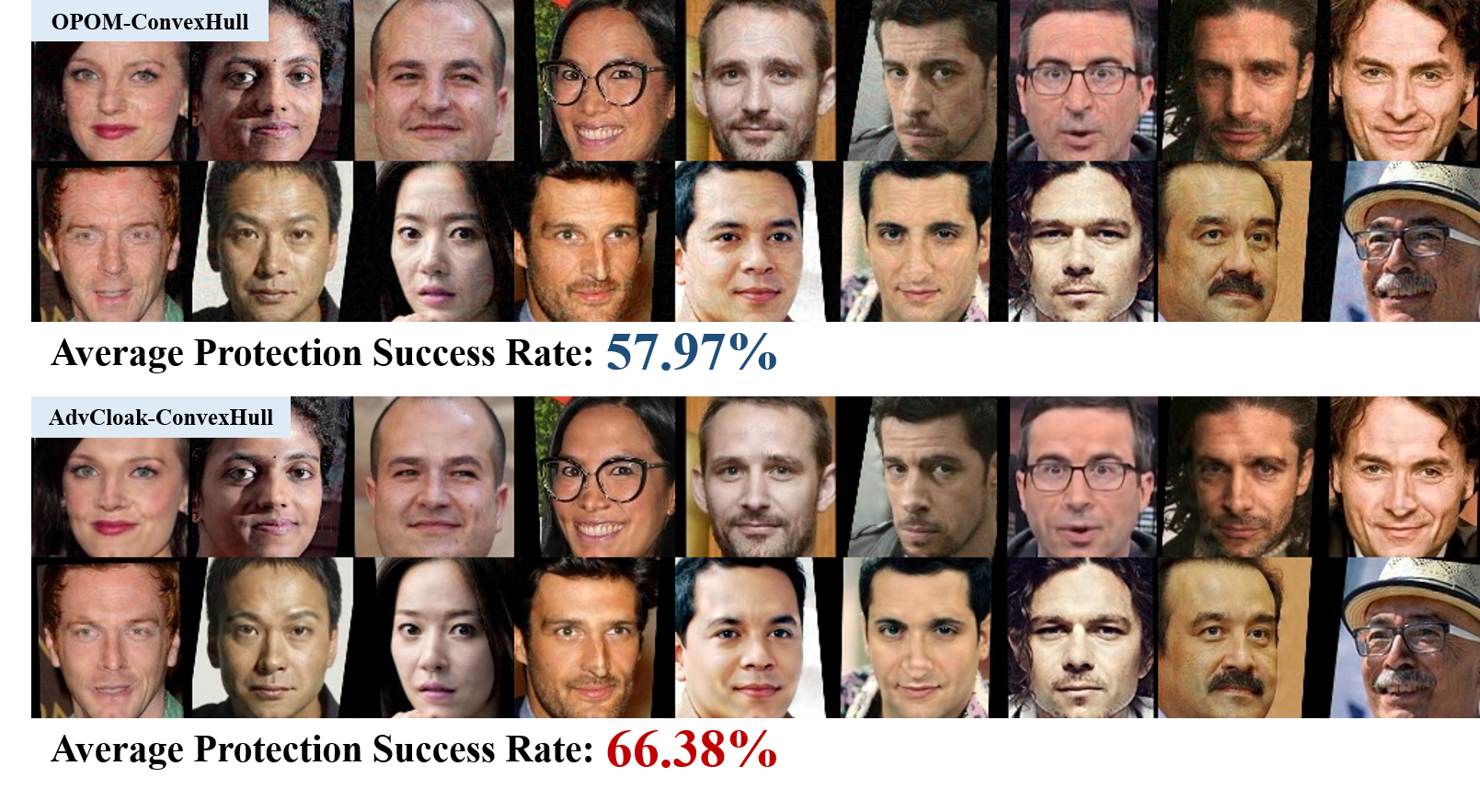}
	\caption{Visualizing the protected adversarial test images in Privacy-Celebrities, generated by OPOM-ConvexHull~\cite{zhong2022opom} and AdvCloak-ConvexHull respectively.} 
	\label{adv_img}
\end{figure*}

\begin{figure*}[!ht]
	\centering
	\includegraphics[width=0.95\linewidth]{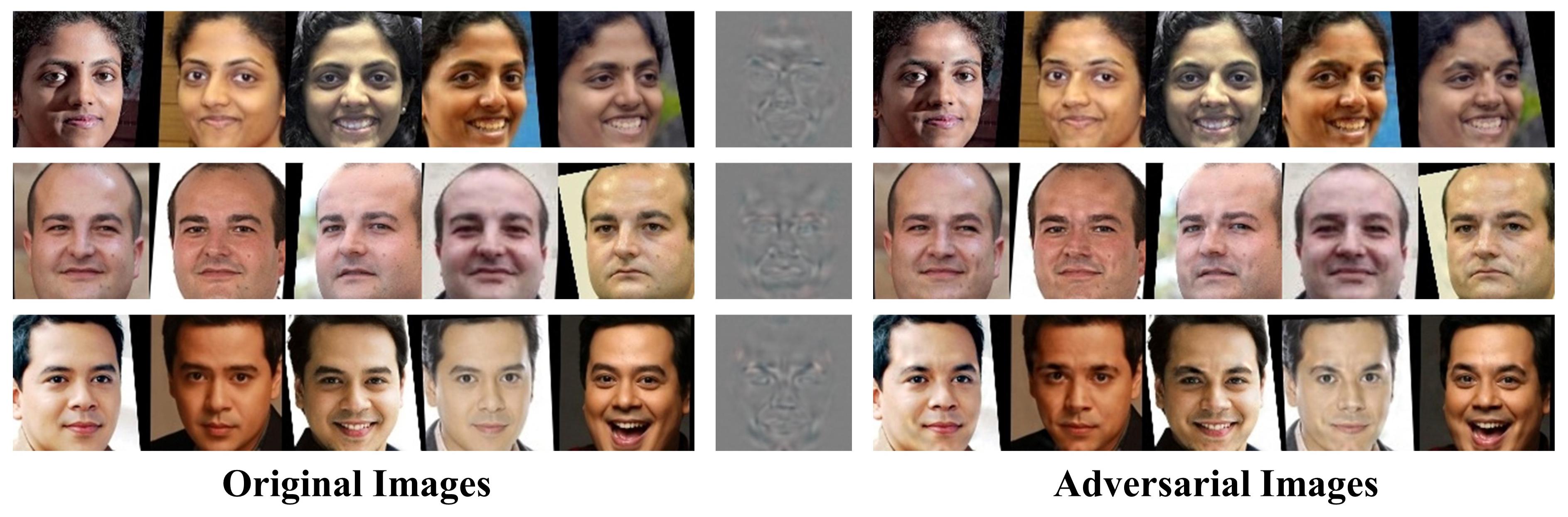}
	\caption{The original images and adversarial images (the protected images) in test datasets of Privacy-Celebrities. The left five columns show original images belonging to three people. The middle column shows the person-specific (class-wise) mask of each person generated by AdvCloak-ConvexHull. The right five columns show the corresponding adversarial images of the same people.} 
	\label{ori_adv}
\end{figure*}

\subsection{Discussion}
\subsubsection{Evaluations on Image Quality of Adversarial Images}
An intuitive idea is that higher perturbation magnitudes usually bring better protection performance at the cost of lower image quality. However, ensuring the naturalness of the generated adversarial images (protected images) is also critical in the task of privacy protection. By scaling the perturbation magnitude of the adversarial masks, we discover the association between the average protection success rate and structural similarity (SSIM) of the universal privacy masks generated by different approaches, as shown in Figure \ref{discuss} (a). We can observe that when compared to alternative methods, our approaches achieve superior image quality at equivalent protection performance, and vice versa. Notably, AdvCloak-ConvexHull surpasses AdvCloak in both structural similarity and average protection success rate under the same perturbation magnitude constraints, demonstrating that feature subspace modeling can simultaneously improve image quality and protection performance.

Figure \ref{adv_img} visualizes some adversarial test images generated by our methods and OPOM-ConvexHull~\cite{zhong2022opom}, showing that our methods achieve both better image quality and higher protection performance. Figure \ref{ori_adv} shows that the person-specific masks generated by AdvCloak-ConvexHull can well generalize to unknown images of the source identity at the image level. 

\subsection{Providing with different numbers of inference images}
Person-specific methods aim to exploit a small number of face images to generate universal adversarial masks. Therefore, we evaluate the relationship between the number of inference images pre-provided and privacy protection performance. In Figure \ref{discuss}(b), we compared our proposed AdvCloak, AdvCloak-ConvexHull with other person-specific methods, \textit{i.e.}, AdvFaces+, FI-UAP~\cite{zhong2022opom}, and OPOM-ConvexHull~\cite{zhong2022opom}, on the Privacy-Commons dataset against the ArcFace model. We utilize the average protection success rate of six black-box models for brevity. The results show that regardless of how many inference images are provided, the AdvCloak-ClassCenter and AdvCloak-ConvexHull outperform other methods in terms of protection performance.

\begin{table}[!h]
	\renewcommand{\arraystretch}{1.3}
	\caption{\centering The evaluation of the efficiency and effectiveness of different masks. $t_1$ represents the time to customize a person-specific (class-wise) mask for each person, and $t_2$ represents the time to generate a corresponding image-specific mask for each face image.}
	\centering
	\vspace{-2mm}
	\label{time}
	\resizebox{\linewidth}{!}{
		\begin{tabular}{c|c|cc|c}
			\toprule
			\multirow{2}{*}{\textbf{Type of Mask}}                     & \multirow{2}{*}{\textbf{Methods}} & \multicolumn{2}{c|}{\textbf{Computation Time}} & \multicolumn{1}{l}{\multirow{2}{*}{\textbf{APSR (\%)} $\uparrow$}} \\
			&                          & \textbf{$t_1$(s) $\downarrow$}                 & \textbf{$t_2$(s) $\downarrow$}           & \multicolumn{1}{l}{}                          \\ \midrule \midrule
			\multirow{3}{*}{Image-specific mask}  & FIM~\cite{zhong2019adversarial}                      & /                    & 1.65           & 89.15                                         \\
			& LowKey~\cite{cherepanova2021lowkey}                   & \textbf{/}           & 2.35           & \textbf{89.59}                                         \\
			& AdvFaces~\cite{deb2020advfaces}                 & \textbf{/}           & 0.01           & 88.09                                         \\ \midrule
			\multirow{5}{*}{Person-Specific mask} & FI-UAP                   & 2.81                 & \textbf{0}              & 65.93                                         \\
			& OPOM-ConvexHull~\cite{zhong2022opom}          & 2.81                 & \textbf{0}              & 71.23                                         \\
			& AdvFaces+                & 0.05                 & \textbf{0}              & 73.35                                         \\
			& AdvCloak                 & 0.05                 & \textbf{0}              & 78.38                                         \\
			& AdvCloak-ConvexHull      & 0.05                 & \textbf{0}              & 79.76                                         \\ \midrule
			Universal mask                        & GAP~\cite{poursaeed2018generative}                      & \textbf{0}                    & \textbf{0}              & 27.55                                         \\ \bottomrule
	\end{tabular}}
\end{table}

\subsubsection{The Effectiveness and Efficiency of AdvCloak}
Here, we analyze the benefits of our proposed method AdvCloak and other methods in terms of effectiveness and efficiency. Specifically, we separately calculate the time required to customize the person-specific mask for each person, denoted as $t_1$, and the time required to generate a corresponding mask for each face image denoted as $t_2$.

The experimental results are shown in Table \ref{time}. Firstly, although image-specific masks acquire the best average protection success rate, they need to generate corresponding mask for each face image. In contrast, person-specific masks and universal masks, once generated, render $t_2$ practically negligible, markedly improving the efficiency of privacy protection. Secondly, compared with other person-specific methods, AdvCloak-ConvexHull achieves better protection performance with lower inference time. In addition, the universal method, GAP~\cite{poursaeed2018generative}, seems to be the most efficient. However, its extremely low protection success rate renders it impractical for real-world applications.

\section{CONCLUSION}
In this paper, we propose a generative model-based method, AdvCloak, to customize a type of person-specific (class-wise) cloak for each individual that can be applied to all his or her images. In contrast to previous gradient-based approaches that lack visual semantic information and limit black-box transferability, AdvCloak employs generative models to capture the data distribution and learn the salient adversarial features. To achieve this, we propose a two-stage training strategy, \textit{i.e.}, image-specific and person-specific training, to jointly optimize networks to adapt to different faces and generalize to face variations of the same person successively. To further enhance the intra-class generalization ability of person-specific masks, we combine the AdvCloak with feature subspace modeling by optimizing each feature of the adversarial image in the direction away from the feature subspace of the identity. Experiments on two test benchmarks and six black-box face recognition systems have demonstrated the superiority of our proposed method compared with SOTA methods in terms of effectiveness, efficiency, and image naturalness in privacy protection tasks.

\section*{ACKNOWLEDGMENT}
This work was supported in part by the National Natural Science Foundation of China under Grant No.62276030 and 62236003 and sponsored by National Natural Science Foundation of China (Grant No. 62306041), Beijing Nova Program (Grant No. Z211100002121106, 20230484488).

% \clearpage
% \section*{Author Biographies}
% \textbf{Xuannan Liu} is a graduate student studying for a doctor’s degree in the School of Artificial Intelligence, Beijing University of Posts and Telecommunications (BUPT), Beijing, China. His research interests include face privacy protection and adversarial example.

% \textbf{Weihong Deng} received the B.E. degree in information engineering and the Ph.D. degree in signal and information processing from the Beijing University of Posts and Telecommunications (BUPT), Beijing, China, in 2004 and 2009, respectively. He is currently a professor in School of Artificial Intelligence, BUPT. His research interests include trustworthy biometrics and affective computing, with a particular emphasis in face recognition and expression analysis.
% \clearpage
\bibliographystyle{elsarticle-num} 
\bibliography{egbib}


\end{document}